\documentclass[10pt,twocolumn,letterpaper]{article}

\usepackage[pagenumbers]{cvpr} 

\usepackage{graphicx}
\usepackage{amsmath}
\usepackage{amssymb}
\usepackage{booktabs}
\usepackage{epsfig}
\usepackage{multirow}
\usepackage{dblfloatfix}
\usepackage{float}

\usepackage[pagebackref,breaklinks,colorlinks]{hyperref}

\usepackage[capitalize]{cleveref}
\crefname{section}{Sec.}{Secs.}
\Crefname{section}{Section}{Sections}
\Crefname{table}{Table}{Tables}
\crefname{table}{Tab.}{Tabs.}

\def\cvprPaperID{15} 
\def\confName{CVPR}
\def\confYear{2022}

\newcommand\heigh{4.8cm}
\newcommand\height{5.2cm}

\begin{document}

\title{Recognition of Freely Selected Keypoints on Human Limbs}

\author{Katja Ludwig \hspace{2cm} Daniel Kienzle \hspace{2cm} Rainer Lienhart\\
Machine Learning and Computer Vision Lab, University of Augsburg\\
{\tt\small \{katja.ludwig, daniel.kienzle, rainer.lienhart\}@uni-a.de}
}

\maketitle

\begin{abstract}
Nearly all Human Pose Estimation (HPE) datasets consist of a fixed set of keypoints. Standard HPE models trained on such datasets can only detect these keypoints. If more points are desired, they have to be manually annotated and the model needs to be retrained. Our approach leverages the Vision Transformer architecture to extend the capability of the model to detect arbitrary keypoints on the limbs of persons. We propose two different approaches to encode the desired keypoints. (1) Each keypoint is defined by its position along the line between the two enclosing keypoints from the fixed set and its relative distance between this line and the edge of the limb. (2) Keypoints are defined as coordinates on a norm pose. Both approaches are based on the TokenPose \cite{tokenpose} architecture, while the keypoint tokens that correspond to the fixed keypoints are replaced with our novel module. Experiments show that our approaches achieve similar results to TokenPose on the fixed keypoints and are capable of detecting arbitrary keypoints on the limbs.
\end{abstract}

\section{Introduction}

Athletes of various sports disciplines use video analysis in order to evaluate their performance and to improve their capabilities based on the results. In team sports, the trajectories of the athletes and e.g. the ball are often at the center of interest. In contrast, in individual sports, the analyses involve mainly the precision and speed of movements of the individual athlete. Therefore, these analyses are often based on the location of keypoints and body parts of the athlete in the video. Triple and long jump athletes, for example, use the keypoint locations to calculate their step frequency and analyze their body posture. 
\begin{figure}[t]
  \centering
  \begin{subfigure}{0.495\linewidth}
\centering    
    \includegraphics[height=4cm]{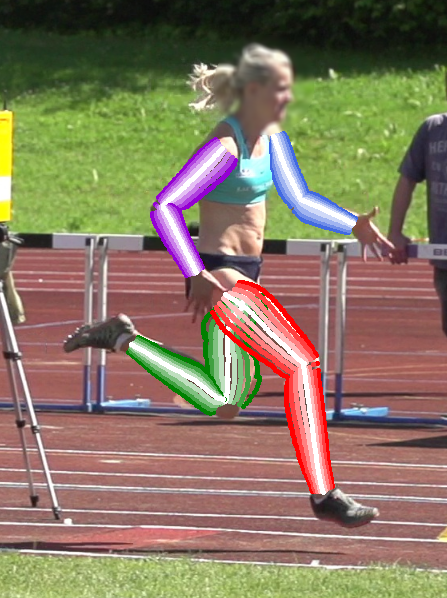}
  \end{subfigure}
  \hfill
  \begin{subfigure}{0.495\linewidth}
  \centering
    \includegraphics[height=4cm]{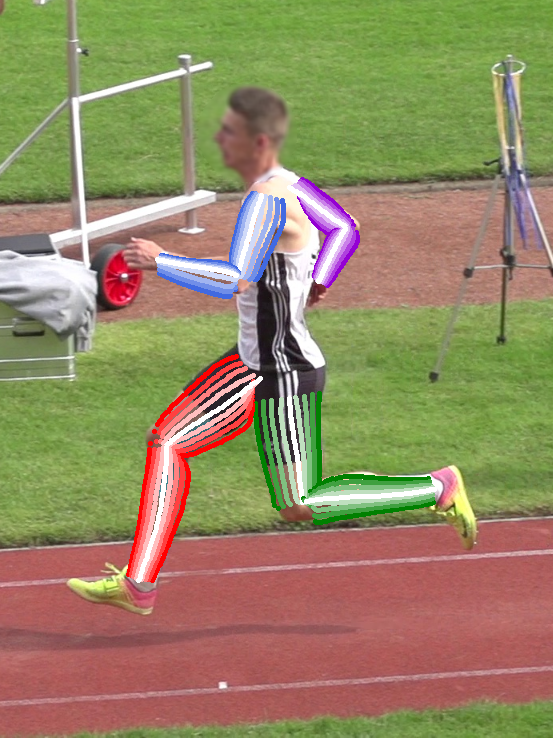}
  \end{subfigure}
   \caption{Two examples of detection results for freely chosen keypoints on the limbs of triple and long jump athletes. The images show four equally spaced lines to both sides of each limb including the edge in pure color and the central line in white.}
   \label{fig:example_prediction}
   \vspace{-0.4cm}
\end{figure}

2D human pose estimation (HPE) techniques can automate the detection of keypoint locations, which makes the video analysis less time consuming and available to more athletes. As annotating images is very time consuming, the datasets of specific sports disciplines are usually small and contain only those keypoints that are essential for the analyses. Other keypoints, for example on the limb boundaries, might open the possibility for new and/or extended types of analyses, but are too expensive to annotate. With our approach, such keypoints can be estimated without any additional annotations.

In computer vision research, 2D HPE  is a task of high interest. The typical problem is to detect a fixed set of keypoints in images of persons. The keypoints have a fixed definition that does not change throughout the task. The goal is to find a model that detects these keypoints as accurately as possible. Commonly, architectures involving deep convolutional neural networks (CNNs) are used. The reason is their high performance in visual tasks. CNNs extract features in a backbone network and predict the keypoint locations in a head module which is specific for the fixed keypoints. Adding new keypoints requires a different head and a retraining of the model. Recently, Transformer \cite{transformer} architectures have become popular among vision tasks. They originate from natural language processing tasks and are designed to handle inputs of various length like sentences. An adaption to vision tasks is achieved by the Vision Transformer \cite{visiontransformer} architecture, which splits images into patches that are handled like words in the original Transformer. Both image patches and words are embedded to vectors of a fixed size in a first step, called tokens.

For HPE, the TokenPose \cite{tokenpose} architecture appends additional learnable tokens to the sequence of tokens from the image patches. 
Our approach uses the capability of Transformer architectures to handle inputs of various length. Hence, we are able to detect the fixed keypoints as well as freely selected keypoints on the human limbs in one step, without the necessity of any costly additional annotations. The representations of the desired keypoints - fixed as well as freely chosen on the limbs - are converted to tokens. This sequence of tokens of arbitrary length is then appended to the image tokens and the network predicts a keypoint for each token. The tokens are generated from keypoint representations. We propose and evaluate two different keypoint representations. The first approach splits the representation into two parts. One part encodes the position of the projection of the desired keypoint onto the straight line between the fixed keypoints that enclose the corresponding body part. The second part encodes the distance of the keypoint from this projection point relative to the distance of the boundary of the body part. We refer to this approach as the \emph{vectorized keypoint} approach. The second approach encodes each keypoint in normalized euclidean coordinates on a norm pose (template pose). We call this the \emph{norm pose} approach. Both approaches open the possibility to design the keypoint representation such that desired arbitrary points on the limbs can be represented and therefore also detected by our model without any additional annotations or postprocessing steps. Figure \ref{fig:example_prediction} shows two examples for such detection results. 
The contributions of this work are as follows:
\begin{itemize}
\item We propose two different representations of freely chosen keypoints on human limbs. The first one is based on the location relative to the body part boundary and the keypoints enclosing the body part, the second one uses the position on a norm pose.
\item Our model, based on the TokenPose architecture, uses the representations to create appropriate tokens for detecting the desired  keypoints. The model can deal with any number of keypoints.
\item We propose a metric to evaluate the location of detected keypoints relative to the body part boundary. Typical metrics like Percentage of Correct Keypoints (PCK) are too inaccurate to evaluate the model's sense of limb boundaries precisely.
\item Our experiments show that the proposed approach can detect arbitrary keypoints on the limbs of humans while maintaining its performance on the set of fixed keypoints. We evaluate the model on the COCO \cite{coco} dataset and a dataset of triple and long jump athletes. 
\end{itemize}

\section{Related Work}

In many sports disciplines, computer vision is a beneficial technique to analyze athletes. Kulkarni et al. \cite{kulkarni2021table} use CNNs to estimate athletes' poses and classify table tennis stroke types. Woinoski et al. \cite{woinoski2021swimmer} detect and track swimmers during races, analyze strokes and detect breaths. Einfalt et al. \cite{einfalt2018activity} detect poses of swimmers and improve their estimated poses by using the swimming style as an additional input to the neural network and pose refinement over time. Moreover, computer vision is also used in team sports. E.g., Bridgeman et al. \cite{bridgeman2019multi} track athletes in soccer videos and create 3D poses of them, while Wei et al.\cite{wei2015predicting} estimate the location of the ball from monocular basketball video footage based on the players' trajectories. Furthermore, human pose and ski estimation is used for different ski disciplines. Wang et al. \cite{wang2019ai} estimate the poses of freestyle skiers and propose a pose correction and exemplar-based visual suggestions to athletes. Further, robust estimation methods for human and ski pose recognition are proposed by Ludwig et al. \cite{ludwig2020robust} in order to calculate the flight angles of ski jumpers during their flight phase.

In sports, 2D HPE  is a very common technique among computer vision analysis applications. The approaches with the best scores on leaderboards of common benchmarks like COCO \cite{coco} or MPII Human Pose \cite{andriluka14cvpr} are based on CNNs \cite{huang2020joint, bulat2020toward}. A common backbone for recent HPE approaches (also used in \cite{huang2020joint}) is the High Resolution Net (HRNet) \cite{hrnet}. It preserves a large resolution throughout the whole network and uses connections between different resolutions instead of an encoder-decoder architecture like in \cite{he2017mask, newell2016stacked, xiao2018simple}. Contrary to the fully convolutional approaches which are most common, TokenPose \cite{tokenpose} is a Transformer \cite{transformer} based approach to HPE. It is usable without any convolutions, but it achieves the best and state-of-the-art results by using the stump of an HRNet as feature extractor. The basic Transformer \cite{transformer} architecture takes sequences of 1D tokens as an input. In order to deal with 2D images or feature maps, Vision Transformer \cite{visiontransformer} proposes to embed small image patches by a learned linear projection to 1D token vectors. This approach is used by TokenPose. Additionally,  learnable keypoint tokens are appended to the image tokens and used as the Transformer input. The output of these keypoint tokens is then transformed through a MLP to heatmaps. This method can be adapted to detect arbitrary keypoints that lie on the straight line between fixed keypoints \cite{ludwig2022detecting}.

However, we are not aware of any related work that addresses the task of estimating freely chosen novel keypoints while training the HPE network only with a training set with fixed keypoint annotations and associated human segmentation masks.

\begin{figure*}[htb]
  \centering
  \begin{subfigure}{0.2\linewidth}
\centering    
    \includegraphics[height=3.7cm]{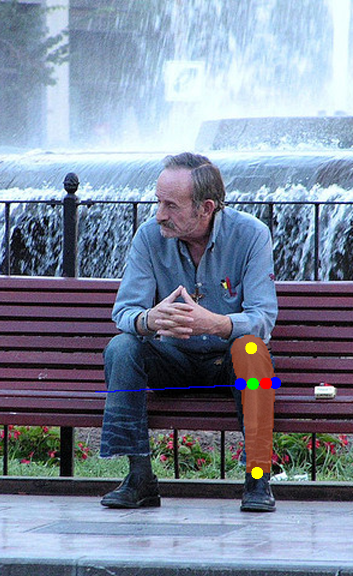}
  \end{subfigure}
  \hfill
  \begin{subfigure}{0.3\linewidth}
  \centering
    \includegraphics[height=3.7cm]{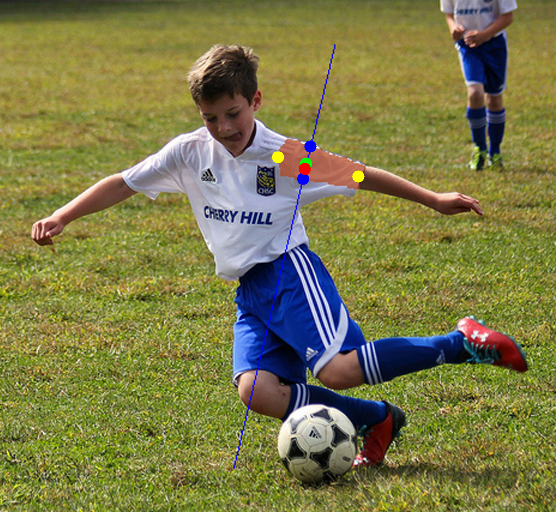}
  \end{subfigure}
  \hfill
  \begin{subfigure}{0.24\linewidth}
  \centering
    \includegraphics[height=3.7cm]{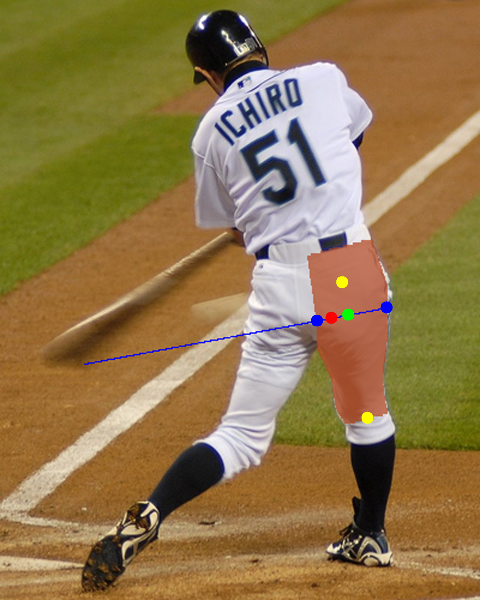}
  \end{subfigure}
  \hfill
  \begin{subfigure}{0.24\linewidth}
  \centering
    \includegraphics[height=3.7cm]{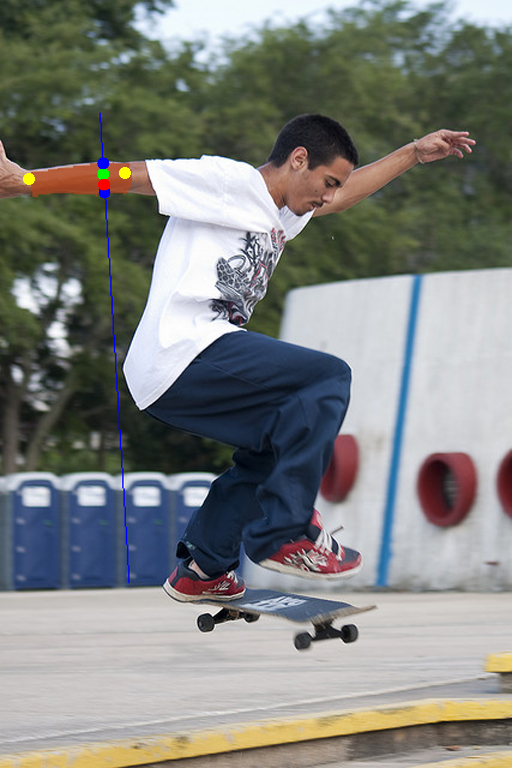}
  \end{subfigure}
  \hfill
  \caption{Examples for the keypoint generation process on COCO images. The body part is visualized with a red overlay and the fixed keypoints enclosing the body part in yellow. The randomly selected projection point on the line between the fixed keypoints is displayed in green and the orthogonal line in blue. The intersection points of the line with the edge of the body part are visualized in blue, while the red points visualize the final generated keypoints.}
  \label{fig:keypoint_generation}
  \vspace{-0.3cm}
\end{figure*}
\section{Method}

The TokenPose-Base \cite{tokenpose} architecture is used as a basis for our model. It uses a convolutional model in the early layers of the backbone and combines it with a Transformer architecture in the later backbone layers. The proposed method can also be used in conjunction with the other 
TokenPose variants.

\subsection{Keypoint Generation}\label{sec:keypoint_gen}

In order to detect arbitrarily selected keypoints on human limbs, we need to generate ground truth keypoints on the limbs. To achieve that, we use segmentation masks of upper arms, forearms, thighs and lower legs. As we want to generate keypoints that are distributed over the complete body part, we use the following generation scheme: Let $b_i$ and $b_j$ be the coordinates of  two fixed keypoints (e.g., left shoulder and left elbow joints) that enclose the body part $B$ (e.g., left upper arm). At first, we uniformly sample a percentage $p_b$ of the line between $b_i$ and $b_j$, which results in the projection point $b_p$:
\begin{equation}
b_p = p_b \cdot b_j + (1 - p_b) \cdot b_i
\end{equation} 
Next, we generate the line $f$ that is orthogonal to the line between $b_i$ and $b_j$ and fits through $b_p$. This line has two intersection points $c_1$ and $c_2$ with the boundary of the body part segmentation mask $B$ . Then, we sample $\tilde{p_t}$ from a normal distribution and define $p_t = \max(0, 1 - |\tilde{p_t}|) \in [0,1]$. This ratio $p_t$ corresponds to the distance from the projection point $b_p$ to the body part boundary, referred to as the \textit{thickness}. With $p_t$, we create the final keypoint $b_t$ as follows:
\begin{equation}
b_t =\left\{\begin{array}{ll} (1- p_t) \cdot b_p + p_t \cdot c_1, & \tilde{p_t} >= 0\\
         								(1 - p_t) \cdot b_p + p_t \cdot c_2, & \tilde{p_t} < 0\end{array}\right .
\end{equation} 

$\tilde{p_t}$ is drawn from a normal distribution in order to generate more keypoints on the body part boundaries, as this seems harder for the model to learn. Figure \ref{fig:keypoint_generation} shows some examples for such keypoint generations. Yellow points visualize $b_i$ and $b_j$, a green point $b_p$ and the blue line $f$. The body part segmentation mask $B$ is visualized by a red overlay. The mask intersection points $c_1$ and $c_2$ are displayed with blue points and the generated point $b_t$ with a red point.

\subsection{Keypoint Representations}

In contrast to TokenPose which uses fixed learnable tokens, we need to learn an embedding function for the desired keypoint to a suitable keypoint token, as it is analyzed in \cite{ludwig2022detecting}. We propose two approaches for the input representation for this embedding function in the following sections.

\subsubsection{Keypoint and Thickness Vectors }\label{sec:vectorized_keypoints}

This approach is directly derived from the keypoint generation process. Each keypoint is represented by two short vectors, a \emph{keypoint vector} and a \emph{thickness vector}. For a dataset with $n$ fixed keypoints, the keypoint vector $v^k \in \mathbb{R}^n$ for the keypoint $b_t$ is designed as follows:

\begin{equation}
v^k_l =\left\{\begin{array}{ll} 1-p_b, & l = i\\
										p_b, & l = j\\
         								0, & l \neq i \land l \neq j\end{array}\right .  l = 1, ..., n
\end{equation}
This is equal to the representation in \cite{ludwig2022detecting} for the projection point $b_p$. The second, novel representation vector is called \emph{thickness vector}, $v^t \in \mathbb{R}^3$, and is defined according to
\begin{equation}
v^t =\left\{\begin{array}{ll} \left(p_t, 1 - p_t, \; 0 \right)^T  , & p_t >= 0\\
										\left(0, \; 1 - p_t,  p_t \right)^T  , & p_t < 0\end{array}\right .  
\end{equation}
The fixed keypoints of the dataset are represented with $p_b = 0$ and $p_t = 0$.

\subsubsection{Norm Pose}\label{sec:norm_pose}
\begin{figure}[htb]
  \centering
  \includegraphics[width=0.5\linewidth]{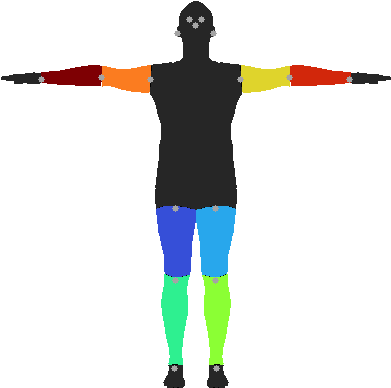}
   \caption{The used norm pose depicted with the fixed keypoints from the COCO dataset.}
   \label{fig:norm_pose}
     \vspace{-0.3cm}
\end{figure}
The norm pose approach encodes the keypoints in normalized 2D-coordinates according to a norm pose. Figure \ref{fig:norm_pose} visualizes the used norm pose. The fixed keypoints from the COCO dataset are displayed in light gray. The used body parts are colored, the rest of the body is visualized in black. The coordinates of the norm pose point $b_n$ are derived in the same way like the keypoint generation for $b_t$ described in Section \ref{sec:keypoint_gen}. The coordinates are normalized to the interval $[0, 1]$. Hence, the norm pose representation is in $\mathbb{R}^2$.

\subsection{Model Architecture}

Our model architecture is closely related to the Tokenpose \cite{tokenpose} architecture, but has important key modifications. Figure \ref{fig:keypoint_vector_architecture} visualizes the general architecture together with the adaption for the keypoint and thickness vectors, which will be explained later. At first, image features are extracted with a CNN. At the beginning of the Transformer, the feature maps are split into equally sized feature patches. The feature patches are embedded to visual tokens by a learned linear projection. A 2D sine positional encoding is added to the visual tokens. Next, the keypoint tokens are appended to this sequence of visual tokens. The creation of these keypoint tokens is dependent on the representation type. We do not add positional encoding to the keypoint tokens as the order of the keypoints should not matter. In the end, a multi-layer perceptron is used to transform the output of the Transformer corresponding to the keypoint tokens to 2D heatmaps.

In a first experiment, called \emph{thickness token} approach in the following, we treat keypoint vectors and thickness vectors similar to feature patches. It transforms the keypoint and thickness vectors to tokens through two independently learned linear projections. Keypoint and thickness tokens are then appended to the visual tokens. The problem with this approach is that the model is not able to match the corresponding keypoint and thickness tokens. Therefore, it predicts the projection points $b_p$ instead of the desired points $b_t$. We would need a positional encoding in order to match the tokens, but this is in contradiction to the desired independence of the order of the keypoints. 

Hence, we use a different approach, which we call \emph{vectorized keypoint} approach. Let $m$ be the desired embedding size of the visual and keypoint tokens. Then, the keypoint vectors and the thickness vectors are embedded to tokens of size $m / 2$ with independently learned linear projections. These tokens are concatenated to the final keypoint tokens of size $m$, which combine the information from keypoint and thickness vectors. These keypoint tokens are appended to the visual tokens and then fed through the Transformer network. An illustration of this model can be found in Figure \ref{fig:keypoint_vector_architecture}. During training, the tokens are first randomly sampled and permuted before being appended to the visual tokens, as described in \cite{ludwig2022detecting}.

\begin{figure}[htb]
  \centering
  \includegraphics[width=0.4\linewidth]{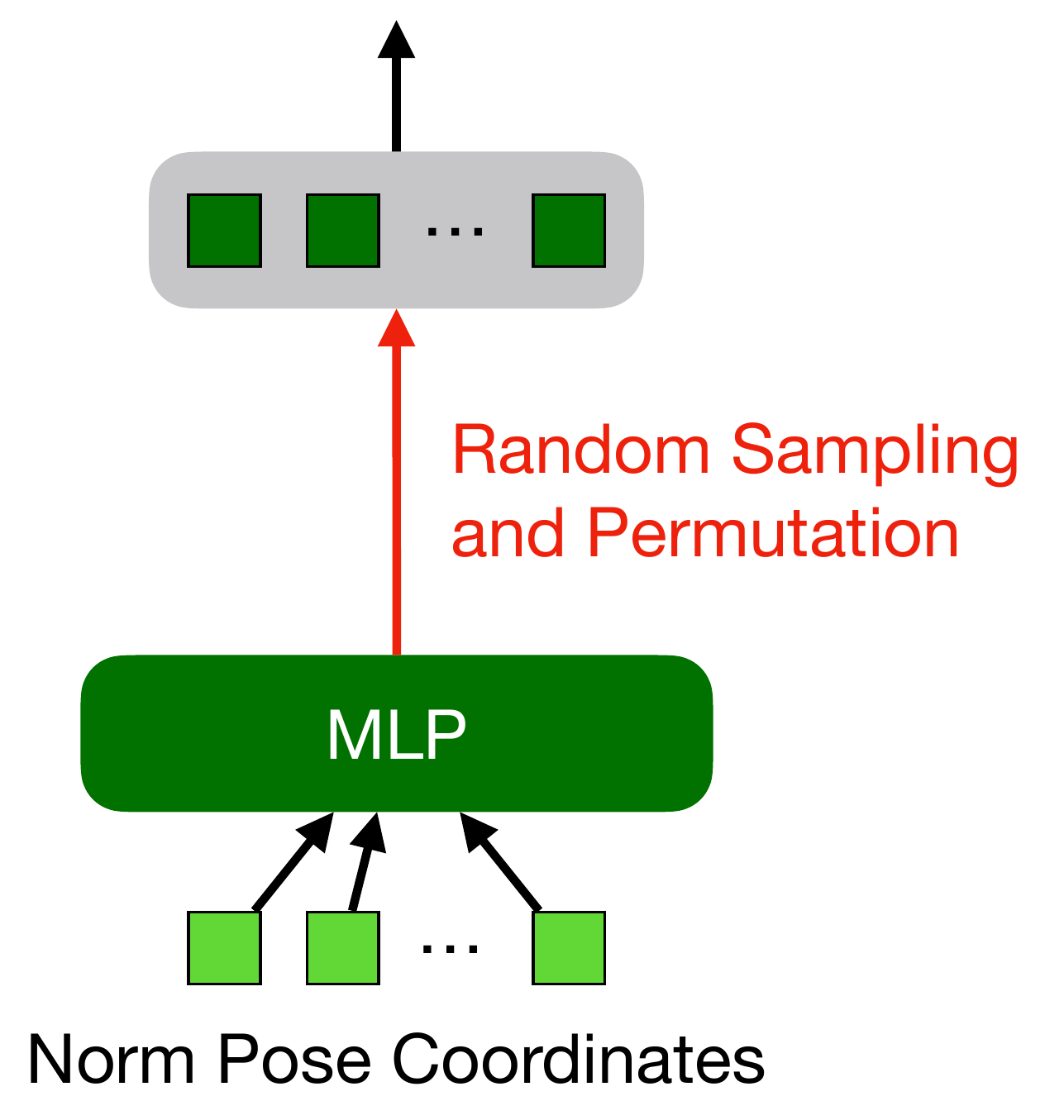}
   \caption{Model architecture adaption for norm pose representations. The norm pose coordinates are transformed to the keypoint vectors via a MLP. Random sampling and permutation applies only during the training phase.}
   \label{fig:norm_pose_architecture}
   \vspace{-0.3cm}
\end{figure}

\begin{figure*}
\begin{center}
\includegraphics[width=0.85\linewidth]{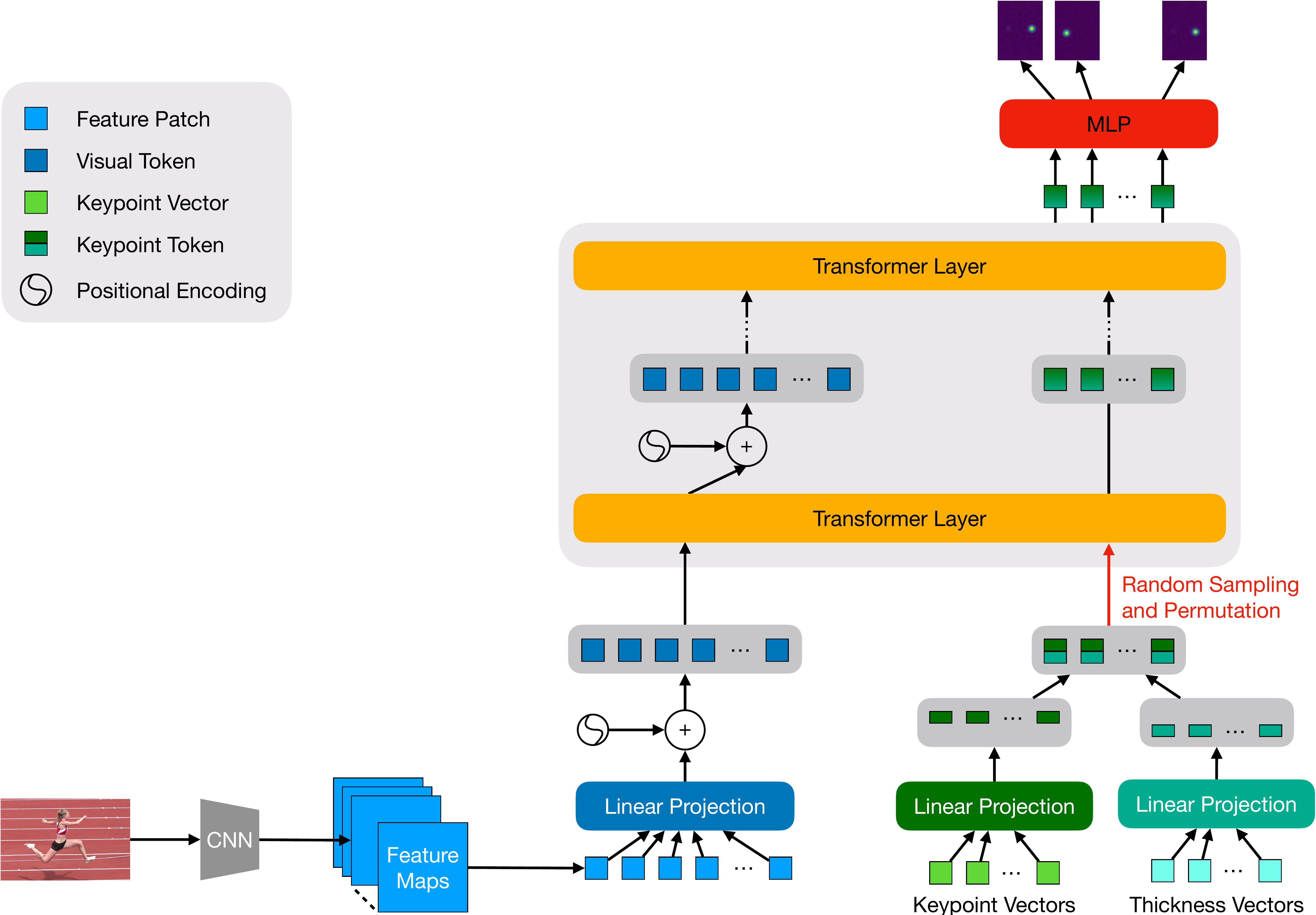}
\end{center}
   \caption{Model architecture with keypoint and thickness vectors. Image features from a CNN are split into patches and transformed to visual tokens via a linear projection. Keypoint and thickness vectors are also embedded via a linear projection, but only to half of the embedding size. Afterwards, they are concatenated to the final keypoint tokens which are appended to the sequence of visual tokens. This sequence is the input to the Transformer network. Random sampling and permutation applies only during the training phase.}
\label{fig:keypoint_vector_architecture}
\end{figure*}

The norm pose coordinates are used similarly. In a first experiment, we embed them as well with a linear projection. However, experiments show that the performance is below the performance of the original TokenPose model. Therefore, we try to enhance the generated keypoint tokens by using a multi-layer perceptron instead of the linear projection in order to give the model more capacity to learn the keypoint semantics. This adaption is visualized in Figure \ref{fig:norm_pose_architecture}. The rest of the model is identical to the model for the vectorized keypoints approach (see Figure 5).
\vspace{-0.2cm}

\subsection{Thickness Metrics}\label{sec:metrics}

Evaluations show that models predicting only the projection points $b_p$ like the thickness token approach achieve significant performance regarding typical metrics like the Percentage of Correct Keypoints (PCK) or the Object Keypoint Similarity (OKS) which are described in Section \ref{sec:experiments}. These metrics are based on the distance between the predicted point and the ground truth point. As the thickness of the limbs is relatively small, the distance between projection points and desired points is also relatively small. This leads to a high performance regarding these metrics, although the model does not learn the semantic of the body part shapes. Therefore, we propose to use a new metric considering the thickness to measure the success of identifying freely selected keypoints correctly.

\begin{figure}[b]
  \centering
  \includegraphics[width=0.42\linewidth]{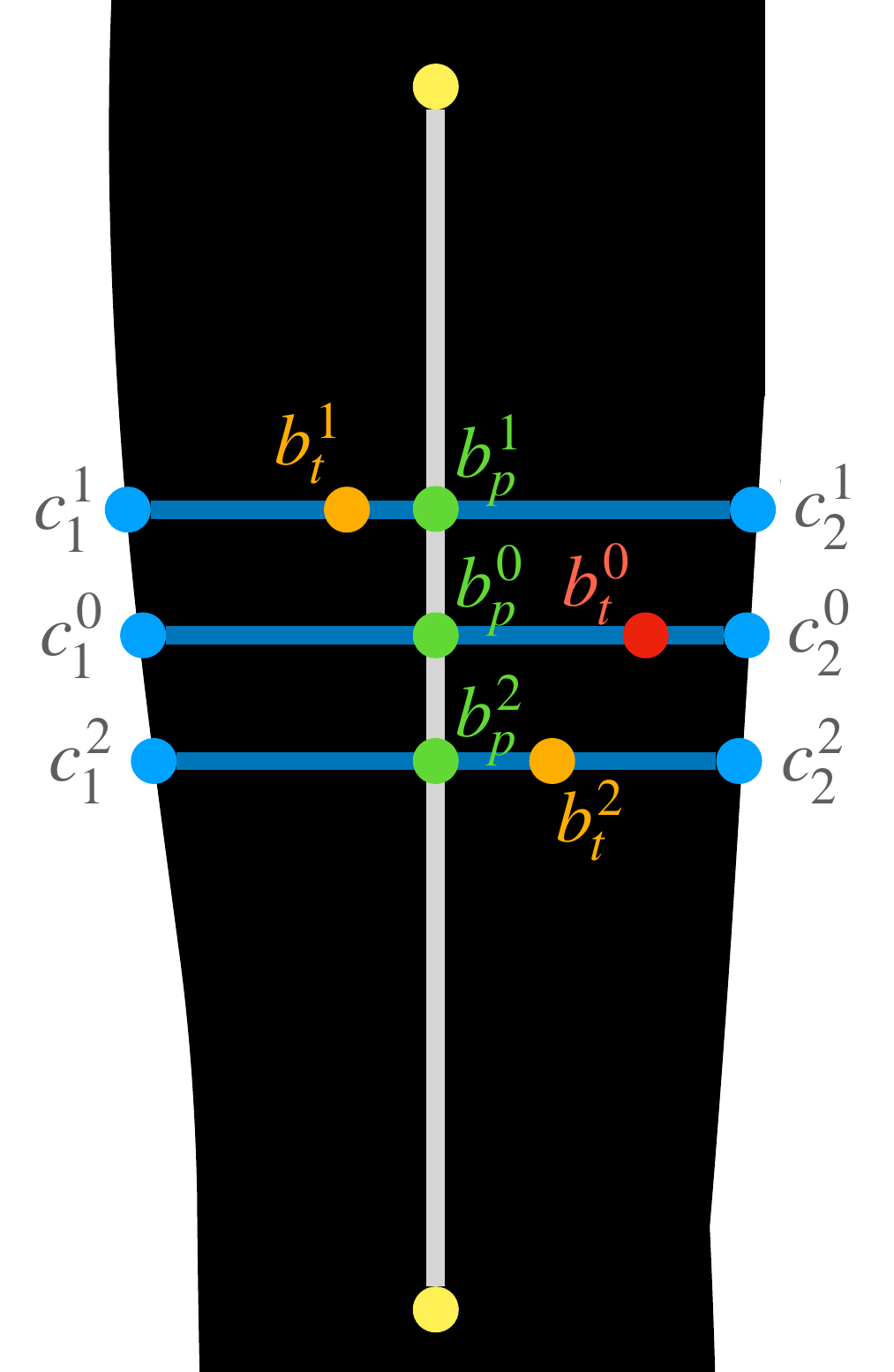}
   \caption{Semantic visualization of the calculation of the thickness error for two possible model predictions. The ground truth is displayed in red, the two predictions in orange. Prediction $b_t^1$ is placed on the opposite side of the gray projection line as the ground truth point $b_t^0$, prediction $b_t^2$ is located on the same side.}
   \label{fig:pct}
\end{figure}

Let $b_t^0$ be the desired ground truth keypoint, $b_p^0$ the corresponding projection point, $c_1^0$ the intersection point on the other side of $b_p^0$ and $c_2^0$ the intersection point on the same side, w.l.o.g., as visualized in Figure \ref{fig:pct}. The ground truth thickness $t_0$
is computed as 
\begin{equation}
t_0 = \frac{||b_t^0 - b_p^0||_2}{||c_2^0 - b_p^0||_2}
\end{equation}
Assume the model predicts a point $b_t^2$ on the same side of the projection line as the ground truth point. Let $b_p^2$ be the projection point corresponding to $b_t^2$ and $c_1^2$, $c_2^2$ be the intersection points in the same way as before. Then, the predicted thickness $t_2$ is calculated as
\begin{equation}
t_2 = \frac{||b_t^2 - b_p^2||_2}{||c_2^2 - b_p^2||_2}
\end{equation}
The thickness error $e_{2}$ is now calculated as the absolute difference between the ground truth thickness and the predicted thickness: $e_{2} = |t_0 - t_2|$.
Furthermore, the model might predict a point $b_t^1$ on the opposite side of the projection line from the ground truth point. With $b_p^1$ being the projection point corresponding to $b_t^1$ and $c_1^1, c_1^2$ be the intersection points on the opposite and same side, respectively, the thickness error $e_1$ is calculated as
\begin{equation}
e_1 = \frac{||b_t^1 - b_p^1||_2}{||c_1^1 - b_p^1||_2} + t_0
\end{equation}
Finally, if a projection point can not be defined for a predicted point $b_t^3$, e.g., because it does not lie in the body part segmentation, we set the thickness error  $e_3$ to the maximum possible thickness error, which is $e_3 = 2$. 
\begin{figure*}[tb]
  \centering
  \begin{subfigure}{0.19\linewidth}
\centering    
    \includegraphics[height=4cm]{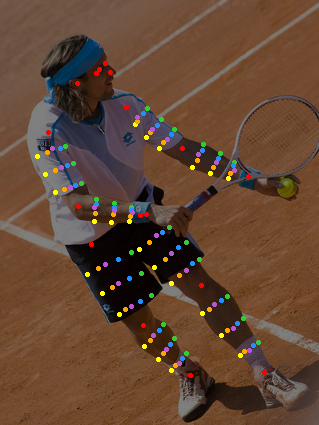}
  \end{subfigure}
  \hfill
  \begin{subfigure}{0.19\linewidth}
\centering    
    \includegraphics[height=4cm]{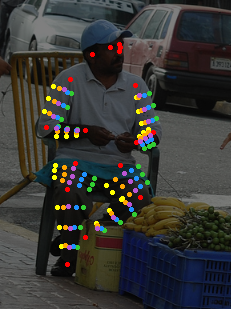}
  \end{subfigure}
  \hfill
  \begin{subfigure}{0.19\linewidth}
\centering    
    \includegraphics[height=4cm]{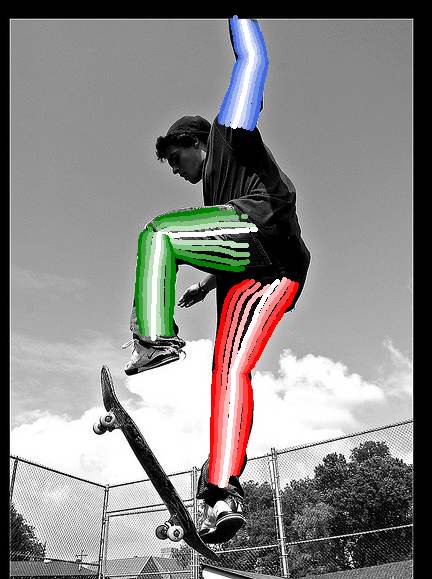}
  \end{subfigure}
  \hfill
  \begin{subfigure}{0.19\linewidth}
	\centering    
    \includegraphics[height=4cm]{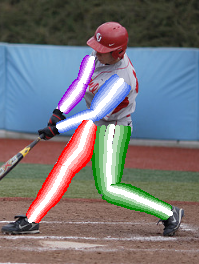}
  \end{subfigure}
  \hfill
  \begin{subfigure}{0.19\linewidth}
	\centering    
    \includegraphics[height=4cm]{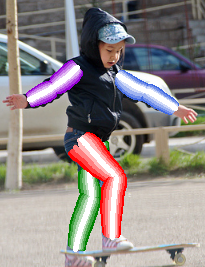}
  \end{subfigure}
  \hfill
  \caption{Examples for model predictions on the DensePose subset of the COCO dataset. The first two images show the fixed keypoints in red and a grid of four equally spaced keypoints along the projection line by five equally spaced keypoints along the thickness line for each body part. The images are darkened for better visibility of the keypoints. The other three images show four equally spaced lines regarding the thickness on each body part. The projection line is colored white with a color gradient to the edges.}
  \label{fig:coco_predictions}
\end{figure*}

As a first metric, we use the \textit{Mean Thickness Error} (MTE). Furthermore, we introduce the \textit{Percentage of Correct Thickness} (PCT). At a threshold $t$, it is defined as the fraction of thickness errors that is below $t$. Notice that these metrics should not be used standalone as they do not take into account the absolute positions of the predictions, only the relative position regarding the projection line and the body part boundaries are considered. They are only able to give a rough estimation, as the thickness error is always set to the maximum error if the keypoint does not lie on the correct body part. Therefore, in our experiments, we use the PCT in conjunction with the PCK.

\section{Experiments}\label{sec:experiments}

All our experiments use the TokenPose-Base \cite{tokenpose} architecture configuration as a backbone. The CNN for feature extraction is an HRNet-w32 \cite{hrnet} pruned to its first three stages. We resize all input images to a size of $256 \times 192$. For the feature patches, we use the largest output feature maps of the HRNet, which are of size $64 \times 48$. These feature maps are split into patches of size $4 \times 3$, which results in 256 feature patches in total. We use 192 as an embedding size, equal to the TokenPose-Base implementation, and 12 Transformer Layers with 8 heads. As positional encoding, we use a 2D sine, which is added only to the visual tokens after the embedding and in between each transformer layer (see Figure \ref{fig:keypoint_vector_architecture}). The MLP after the Transformer layers converts each output corresponding to the keypoint tokens to heatmaps of size $64 \times 48$. The final keypoint coordinates are retrieved with the DARK method \cite{zhang2020distribution}.

\subsection{COCO}\label{sec:coco}
\textbf{Dataset.}
The original COCO \cite{coco} dataset contains over 200,000 images. For our task, we need body part segmentation masks in order to generate arbitrary keypoints on the limbs. Therefore, we use the subset of COCO created for the DensePose \cite{densepose} task. We use the train1 split containing 39.210 person segmentations as our training set, the val split with 2,243 person segmentations as our validation set and the train2 split with 7,297 as our test set. During the keypoint generation process, we found out that the segmentation masks contain a lot of wrong left-right annotations. We corrected some of them with a heuristic and some manually, resulting in approx. 3500 annotation corrections that are publicly available at: \url{https://www.uni-augsburg.de/en/fakultaet/fai/informatik/prof/mmc/research/datensatze/}. The fixed and semantically well-defined keypoints in the COCO dataset are: l./r. eye, l./r. ear, l./r. shoulder, l./r. elbow, l./r. wrist, l./r. hip, l./r. knee, l./r. ankle.

\begin{table*}[htb]
\begin{center}
\begin{tabular}{c|cccccc|cc|cc}
\toprule
Model  & $AP$ & $AP^{50}$ & $AP^{75}$ & $AP^M$ & $AP^L$ & $AR$ & Avg PCK & Full PCK & MTE $\downarrow$ & PCT $\uparrow$\\
\midrule
  TokenPose  &  \textbf{84.6} &  \textbf{97.8} &  \textbf{92.2} &  \textbf{78.9} &  \textbf{85.1} &  \textbf{87.3}  & 84.1\\
 \midrule
  Thickness Tokens &   82.8 &  \textbf{97.8} &  91.0 &  76.9 &  83.3 &  85.8 &   83.0 & 71.0 & 79.2 & 6.3\\ 
  Vectorized Keypoints &  84.0 &  \textbf{97.8} &  92.1 &  78.3 &  84.3 &  86.7 & \textbf{84.2} & \textbf{87.2} & \textbf{25.5} & \textbf{68.1}\\ 
  Norm Pose Linear&   78.5 &  96.7 &  87.6 &  72.7 &  79.1 &  82.1 & 80.5 & 83.1 & 33.0 & 56.4\\
  Norm Pose MLP&    83.1 &  \textbf{97.8}  &  91.2 &  78.0 &  83.6 &  86.0 & 83.7 & 87.1 & 25.7 & 66.9\\
 \bottomrule
\end{tabular}
\end{center}
\vspace{-0.25cm}
\caption{OKS results, PCK@$0.1$ and thickness metrics results on our test set of the DensePose dataset. The Avg PCK is the PCK@$0.1$ metric on the fixed keypoints, the Full PCK the PCK@$0.1$ on the fixed and generated keypoints. MTE and PCT refer to the metrics proposed in Section \ref{sec:metrics}. The TokenPose model is trained only on the fixed keypoints. The \emph{thickness token} approach refers to the model with distinct tokens for thickness and keypoint vectors. The \emph{vectorized keypoint} approach is described in Section \ref{sec:vectorized_keypoints}. \emph{Norm pose MLP} refers to the approach with norm pose representations and a four layer MLP for the embedding, \emph{Norm pose linear} uses a linear projection.}\label{tab:coco_results}
\end{table*}

\textbf{Evaluation Metric.}  The primary metric for keypoint detection on  COCO  is the average precision (AP) based on the Object Keypoint Similarity (OKS). Let $d_i$ be the euclidean distance between corresponding ground truth and detected keypoint, $v_i$ the ground truth visibility flag, $s$ the object scale and $k_i$ a per-keypoint specific constant. The OKS is defined as
\begin{equation}
\mathit{OKS} = \frac{\sum_i \mathit{exp}(-d_i^2 / 2s^2k_i^2) \sigma(v_i>0))}{\sum_i \sigma(v_i>0)}
\end{equation}
The keypoint specific constants are used to control the demanded prediction accuracy based on the keypoint type. As these constants cannot be defined for arbitrary keypoints, we additionally use the PCK metric at threshold 0.1. The PCK@$t$ considers a keypoint prediction correct at a threshold $t$, if the distance between the predicion and the ground truth is less than or equal to $t$ times the torso size. We use the distance between left shoulder and right hip as the torso size. The recall at a certain PCK threshold represents the fraction of keypoints that is considered correct at that threshold. Furthermore, we use the MTE and PCT metric with a threshold of 0.2 as described in Section \ref{sec:metrics} to measure the ability of the model to predict points at the right distance from the projection line. As the maximum error for the PCT metric is 2, we consider $0.2$ as a good threshold for evaluations.

\textbf{Results.}
\begin{figure}[b]
\vspace{-0.3cm}
  \centering
  \includegraphics[width=0.4\linewidth]{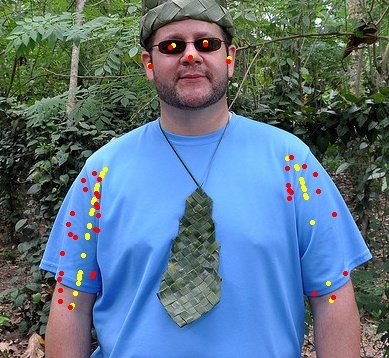}
   \caption{Example predictions for the thickness token model. The predictions displayed in yellow are located only on the projection line and do not consider the thickness of the body parts. This behavior motivates the need for the MTE and PCT metrics. Ground truth keypoints are displayed in red.}
   \label{fig:line_predictions}
\end{figure}
Table \ref{tab:coco_results} displays the results on the DensePose subset of the COCO dataset. The TokenPose baseline approach achieves the best results on the fixed keypoints regarding the AP, but it is not capable of detecting arbitrary points on human limbs. For the other proposed approaches, the focus is shifted from the standard fixed keypoints to the freely selectable points on the limbs, which is the reason for the small decrease in AP for OKS regarding the other approaches. The vectorized keypoint approach achieves a slighly lower AP for OKS on the fixed points, but the PCK for the fixed points is slightly higher and the PCK for all points including the generated keypoints (named \emph{Full PCK} in Table \ref{tab:coco_results}) is even higher (absolute 3.0\%). The full PCK for the approach with independent thickness tokens is lower than the full PCK for the vectorized keypoint model by a large margin of absolute 16.2\%. The reason is that the thickness token approach can not match the thickness tokens to the keypoint tokens as the Transfomer is independent of the order of the input sequence. Figure \ref{fig:line_predictions} shows an example for this problem. 
Many keypoints lie in the distance from the ground truth that is valid for the PCK, therefore the full PCK is still quite high. This is the reason why we propose the consideration of the MTE and the PCT. For the thickness token approach, the mean thickness error is 79.2\%, which is really high compared to the mean thickness error of the vectorized keypoint approach with 25.5\%. The PCT metric makes the difference even clearer. Regarding the vectorized keypoint approach, 68.1\% of the detected keypoints are regarded as correct at a threshold of $0.2$. This is over 10 times better than the PCT achieved by the thickness token approach. 
Furthermore, the norm pose approach with linear embedding achieves the worst results, but using a four layer MLP increases the AP by absolute 4.6\%, which is only absolute 1.5\% below TokenPose on the fixed points. Furthermore, the usage of a MLP improves all other metrics slightly, including the thickness metrics. Overall, the norm pose MLP approach achieves slightly worse, but similar results like the vectorized keypoint approach regarding all metrics. Figure \ref{fig:coco_predictions} shows some qualitative results for the vectorized keypoint approach on the coco dataset.

\subsection{Triple and Long Jump}

\begin{figure*}[htb]
\begin{minipage}[b]{.16\linewidth}
 \centerline{\epsfig{figure=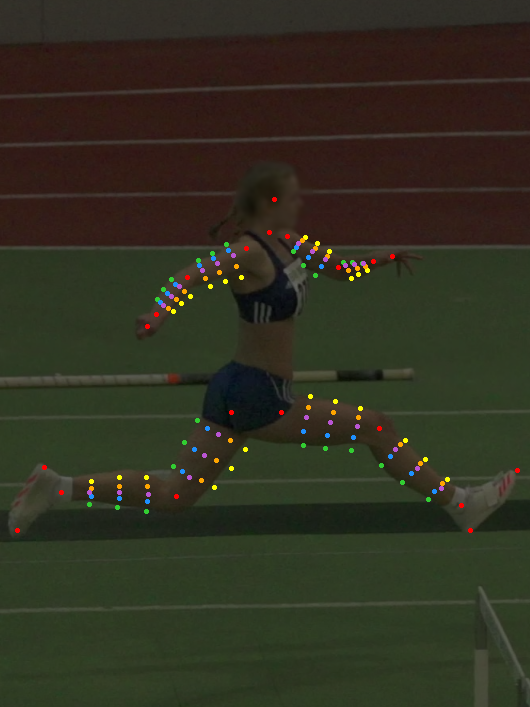,height=3.7cm}}  
\end{minipage}
\hfill
\begin{minipage}[b]{.16\linewidth}
 \centerline{\epsfig{figure=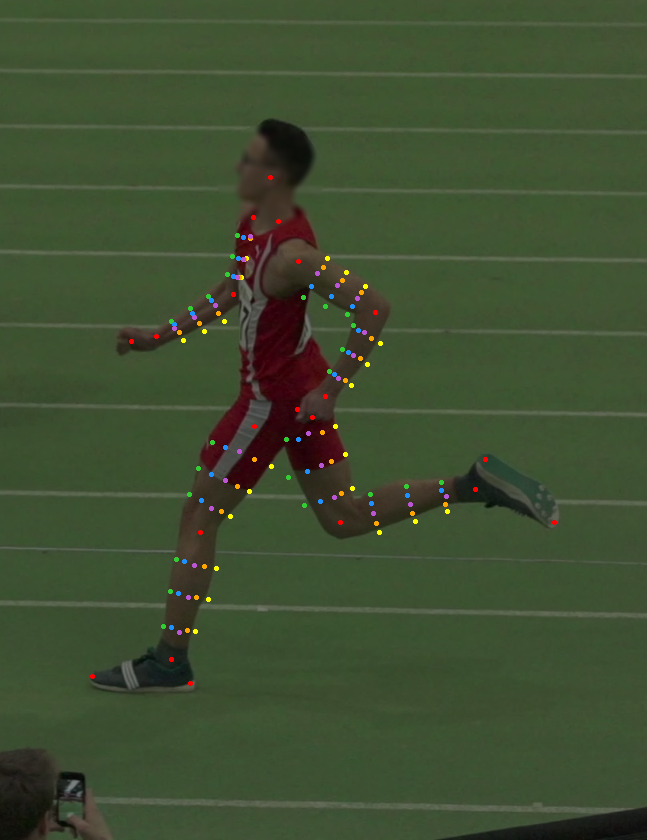,height=3.7cm}}  
\end{minipage}
\hfill
\begin{minipage}[b]{.16\linewidth}
 \centerline{\epsfig{figure=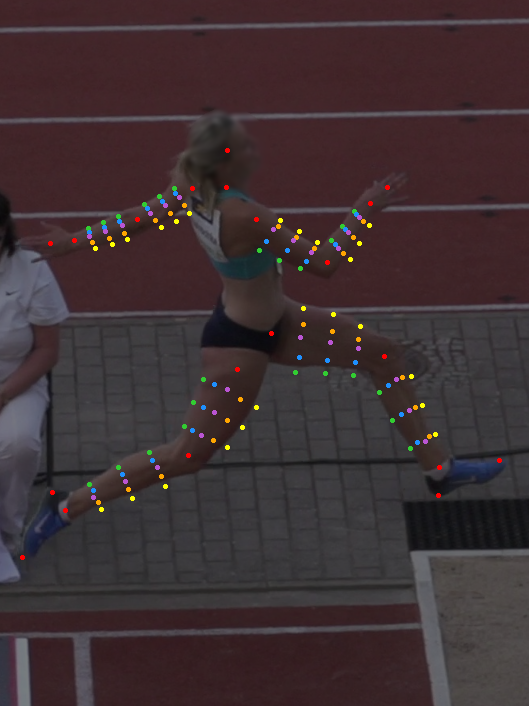,height=3.7cm}}  
\end{minipage}
\hfill
\begin{minipage}[b]{.16\linewidth}
 \centerline{\epsfig{figure=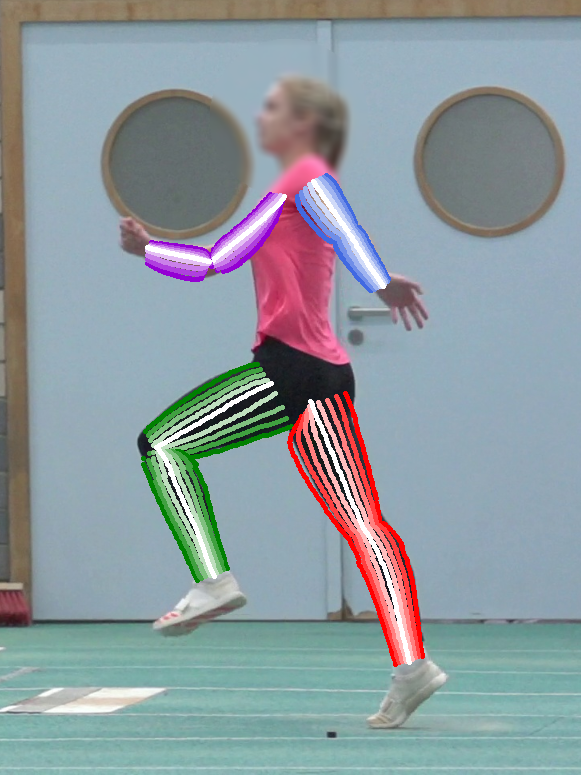,height=3.7cm}}  
\end{minipage}
\hfill
\begin{minipage}[b]{.16\linewidth}
 \centerline{\epsfig{figure=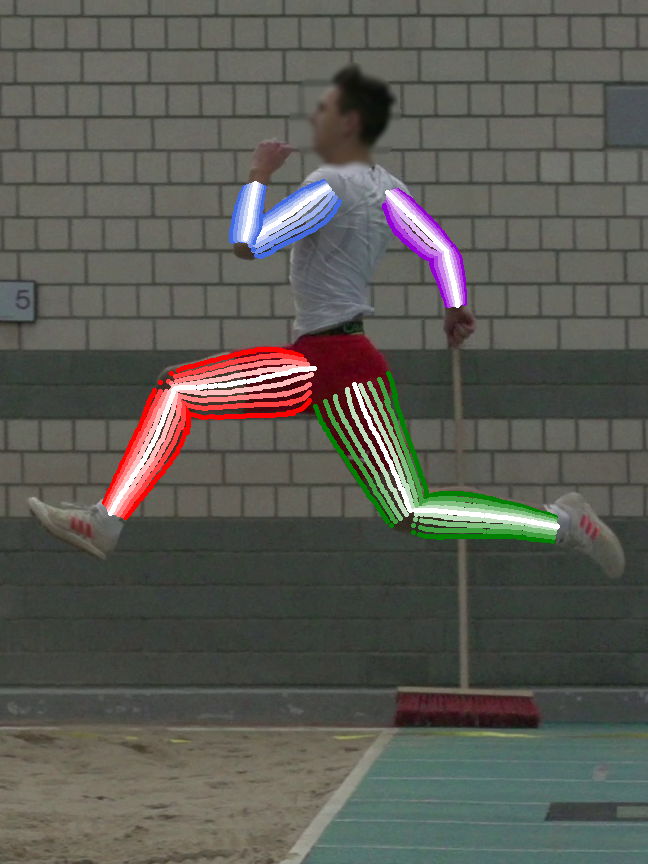,height=3.7cm}}  
\end{minipage}
\hfill
 \begin{minipage}[b]{.16\linewidth}
 \centerline{\epsfig{figure=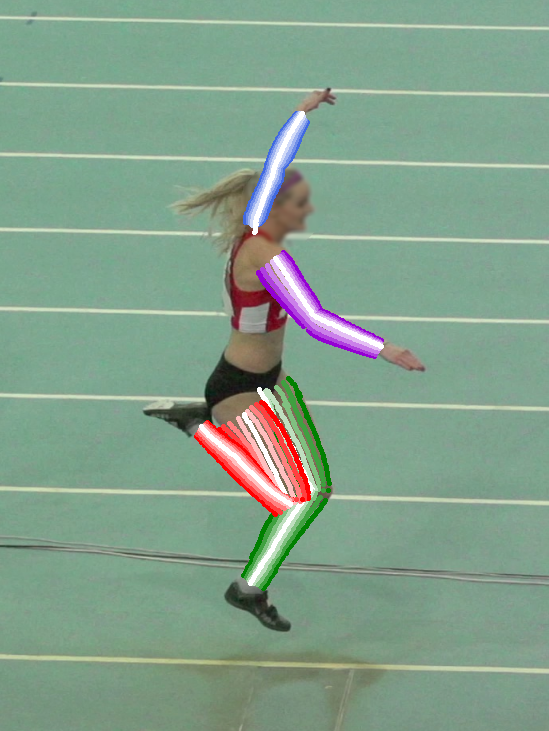,height=3.7cm}}  
\end{minipage}
\vspace{-0.1cm}
\caption{Qualitative results for the triple and long jump test set. The first three images show the fixed keypoints in red and a grid of four equally spaced keypoints along the projection line by five equally spaced keypoints along the thickness line for each body part. The images are darkened for better visibility of the keypoints. The other three images show four equally spaced lines regarding the thickness on each body part. The projection line is colored white with a color gradient to the edges.}
\label{fig:jump_predictions}
\vspace{-0.1cm}
\end{figure*}

\textbf{Dataset.} The triple and long jump dataset consists of frames from videos of triple and long jump athletes during competitions and trainings. The frames show a variety of sports sites and athletes, like indoor and outdoor videos, different lighting conditions, etc. The dataset contains 6,026 labeled images in total, whereby 4,101 images are used for training, 464 images for validation and 1,461 images for the test set. All frames are annotated with head, neck, r./l. shoulder, r./l. elbow, r./l.wrist, r./l. hip, r./l. knee, r./l. ankle, r./l. big toe, r./l. small toe and r./l. heel (20 keypoints in total). The dataset does not contain body part segmentation masks. Therefore, we use the DensePose \cite{densepose} model with a ResNet101 \cite{resnet} backbone and DeepLabV3 \cite{deeplabv3} as well as Panoptic FPN \cite{panopticfpn} heads from detectron2 \cite{densepose} to generate them. Hence, there is no need to costly annotate sports datasets with body part segmentation masks in order to use our method.

\textbf{Evaluation Metric.} We use again the PCK metric as described in Section \ref{sec:coco} with the distance between left shoulder and right hip as the torso size. Like before, we use $t=0.1$, which corresponds to approx. 6\,cm in this dataset. Additionally, we use the MTE and the PCT at a threshold of $0.2$ to evaluate the thickness of the model's predictions.

\textbf{Results.}
The results for the jump dataset are similar to the COCO results and displayed in Table \ref{tab:jump_results}. In comparison to the TokenPose model trained on the fixed keypoints, the vectorized keypoint and the norm pose approach achieve absolute 0.4\% lower PCK on the fixed keypoints, but absolute 2.6\% higher PCK if the generated arbitrary keypoints on the limbs are also considered. Compared to the DensePose COCO dataset, the vectorized keypoint model achieves better results regarding the thickness of the limbs. The MTE is a third lower and the PCT is also a lot higher, absolute 13.3\%. Furthermore, the difference in the performance between linear and MLP norm pose approaches is lower. The vectorized keypoint approach also achieves the best results on this dataset, but the difference to the norm pose MLP approach is only marginally.
In addition, Figure \ref{fig:jump_predictions} visualizes some qualitative results for the jump dataset, which prove that the model has learned a sense of thickness.
\begin{table}[hb]
\begin{center}
\resizebox{\linewidth}{!}{ 
\begin{tabular}{c|cc|cc}
\toprule
Model & Avg PCK  & Full PCK & MTE $\downarrow$ & PCT $\uparrow$\\
\midrule
 TokenPose & \textbf{91.3}\\ 
 \midrule
  Vectorized Keypoints &  \multirow{1}{*}{90.9} &  \multirow{1}{*}{\textbf{93.6}} &  \multirow{1}{*}{\textbf{16.2}} &  \multirow{1}{*}{\textbf{81.4}}\\
  Norm Pose Linear &  90.3 & 93.5 & 17.0 & 79.0\\
  Norm Pose MLP & 90.9 & \textbf{93.6} & 16.8 & 79.8 \\
 \bottomrule
\end{tabular}
}
\end{center}
\vspace{-0.3cm}
\caption{Recall values for the triple and long jump test set in \% at PCK@$0.1$. The first column displays the average PCK of the standard keypoints. The average PCK score including the generated points is given in the second column. The third column shows the MTE and the last column the PCT at threshold $0.1$. The TokenPose model is trained only on the fixed keypoints.} \label{tab:jump_results}
\end{table}
\vspace{-0.3cm}
\section{Conclusion}
This paper proposes two representations for  freely selectable keypoints on the limbs of humans. The first approach, called \emph{vectorized keypoints}, represents each keypoint as a combination of the projection point encoded in a keypoint vector and the thickness encoded in a thickness vector. The projection point is the point on the line between the two fixed keypoints that enclose the body part, while the thickness indicates the distance of the desired keypoint from the projection point to the body part boundary. The \emph{norm pose MLP} approach encodes the desired keypoint as normalized 2D-coordinates relative to a norm pose and uses a small MLP for the embedding to keypoint tokens. In order to evaluate the ability of the model to detect keypoints with the correct thickness, we propose to use the Mean Thickness Error (MTE) and the Percentage of Correct Thickness (PCT) analogous to the PCK metric.

Embedding both keypoint and thickness vectors independently and adding the resulting two tokens to the transformer input sequence leads to the problem that the model detects only keypoints on the line between the enclosing fixed keypoints. This is captured by low PCT scores, despite the quite high PCK and AP of the OKS metric. This proves the necessity of the PCT metric. 
The norm pose approach, if the norm pose is embedded not only with a linear layer but with a MLP, achieves satisfactory results on both datasets. But in comparison to the vectorized keypoint approach, it performs slightly worse on all metrics. Quantitative and qualitative evaluations show that both proposed approaches can successfully detect arbitrary points on the limbs of humans. They achieves high PCT scores, low MTE values while maintaining high PCK (and OKS) scores on both the DensePose subset of the COCO dataset and the triple and long jump dataset.
In the future, we plan to extend our model to arbitrary points anywhere on the human body and not just on the limbs.

\section{Acknowledgements}

This work was funded by the Federal Institute for Sports Science (BISp) based on a resolution of the German Bundestag. We would like to thank the Olympic Training Center Hessen for collecting and providing the triple and long jump data.

{\small
\bibliographystyle{ieee_fullname}
\bibliography{egbib}
}
\twocolumn[{
\makeatletter
\begin{center}
      {\Large \bf Recognition of Freely Selected Keypoints on Human Limbs \\\large Supplementary Material\par}
      \vspace*{24pt}
      
      \large
      \lineskip .5em
      \begin{tabular}[t]{c}
          Katja Ludwig \hspace{2cm} Daniel Kienzle \hspace{2cm} Rainer Lienhart\\
Machine Learning and Computer Vision Lab, University of Augsburg\\
{\tt\small \{katja.ludwig, daniel.kienzle, rainer.lienhart\}@uni-a.de}
      \end{tabular}
      \par
      
      \vskip .5em
      \vspace*{12pt}
   \end{center}
\makeatother}]
\setcounter{figure}{0} 
\renewcommand{\thefigure}{S.\arabic{figure}}

\begin{minipage}{1.0\textwidth}
  \begin{minipage}{0.24\linewidth}  
  \centering
    \includegraphics[height=\heigh]{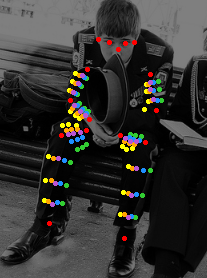}
  \end{minipage}  \vspace{0.1cm}
  \hfill
  \begin{minipage}{0.24\linewidth}
    \centering
    \includegraphics[height=\heigh]{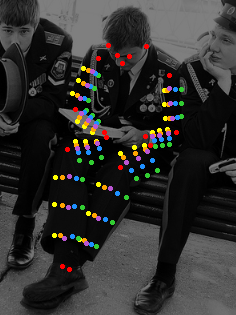}
  \end{minipage}
  \hfill
  \begin{minipage}{0.24\linewidth}
    \centering
    \includegraphics[height=\heigh]{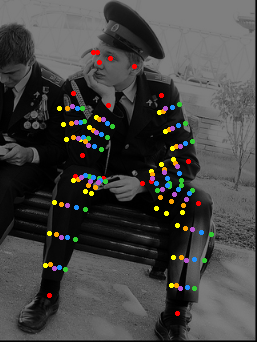}
  \end{minipage}
  \hfill
  \begin{minipage}{0.24\linewidth}
    \centering
      \includegraphics[height=\heigh]{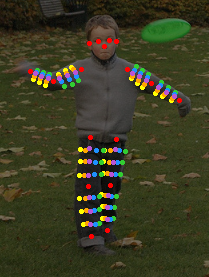}
  \end{minipage}
  \hfill
  \\
  \begin{minipage}{0.24\linewidth}    
    \centering
    \includegraphics[height=\heigh]{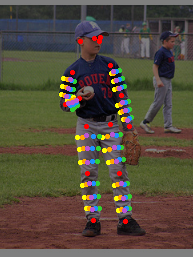}
  \end{minipage} \vspace{0.1cm}
  \hfill 
  \begin{minipage}{0.24\linewidth}  
    \centering
    \includegraphics[height=\heigh]{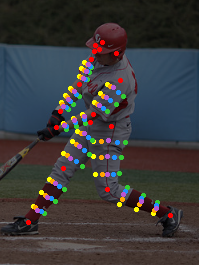}
  \end{minipage}
  \hfill  
  \begin{minipage}{0.24\linewidth} 
    \centering
    \includegraphics[height=\heigh]{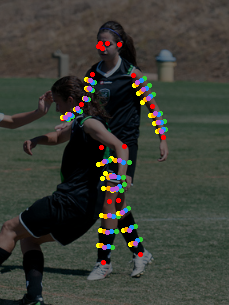}
  \end{minipage}
  \hfill  
  \begin{minipage}{0.24\linewidth}  
    \centering
    \includegraphics[height=\heigh]{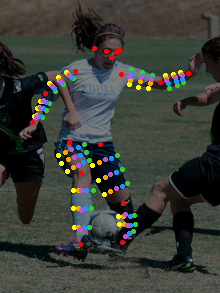}
  \end{minipage}
  \hfill
    \\
  \begin{minipage}{0.24\linewidth}    
    \centering
    \includegraphics[height=\heigh]{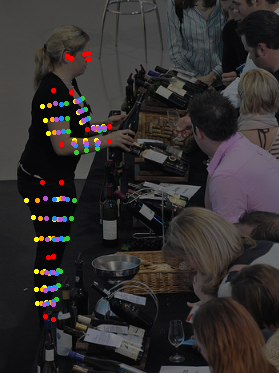}
  \end{minipage} 
  \hfill 
  \begin{minipage}{0.24\linewidth}  
    \centering
    \includegraphics[height=\heigh]{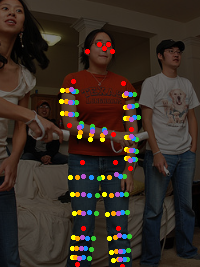}
  \end{minipage}
  \hfill  
  \begin{minipage}{0.24\linewidth} 
    \centering
    \includegraphics[height=\heigh]{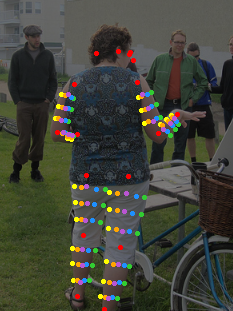}
  \end{minipage}
  \hfill  
  \begin{minipage}{0.24\linewidth}  
    \centering
    \includegraphics[height=\heigh]{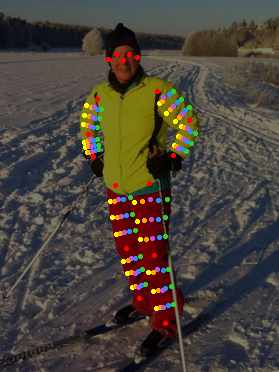}
  \end{minipage}
  \hfill
  \captionof{figure}{Examples for model predictions on the DensePose subset of the COCO dataset. The images show the fixed keypoints in red and a grid of four equally spaced keypoints along the projection line by five equally spaced keypoints along the thickness line for each body part. The images are darkened for better visibility of the keypoints.}
\end{minipage}

\begin{figure*}[ht]
  \begin{subfigure}{0.24\linewidth}  
  \centering
    \includegraphics[height=\height]{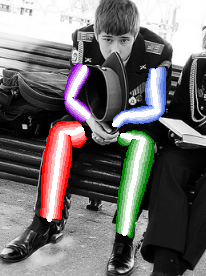}
  \end{subfigure}  \vspace{0.1cm}
  \hfill
  \begin{subfigure}{0.24\linewidth}
    \centering
    \includegraphics[height=\height]{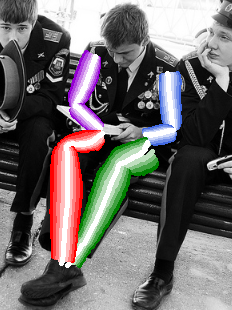}
  \end{subfigure}
  \hfill
  \begin{subfigure}{0.24\linewidth}
    \centering
    \includegraphics[height=\height]{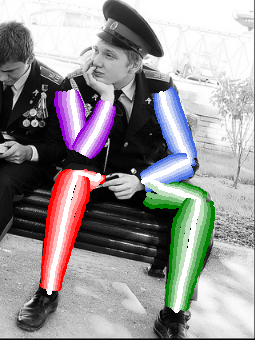}
  \end{subfigure}
  \hfill
  \begin{subfigure}{0.24\linewidth}
    \centering
      \includegraphics[height=\height]{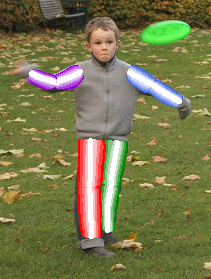}
  \end{subfigure}
  \hfill
  \\
  \begin{subfigure}{0.24\linewidth}    
    \centering
    \includegraphics[height=\height]{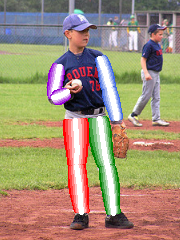}
  \end{subfigure} \vspace{0.1cm}
  \hfill 
  \begin{subfigure}{0.24\linewidth}  
    \centering
    \includegraphics[height=\height]{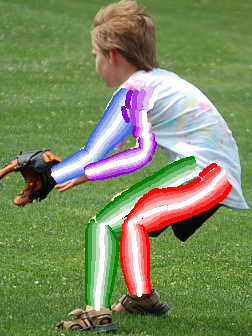}
  \end{subfigure}
  \hfill  
  \begin{subfigure}{0.24\linewidth} 
    \centering
    \includegraphics[height=\height]{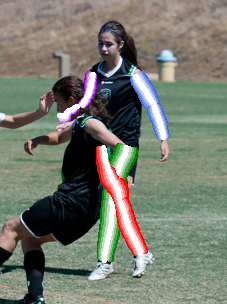}
  \end{subfigure}
  \hfill  
  \begin{subfigure}{0.24\linewidth}  
    \centering
    \includegraphics[height=\height]{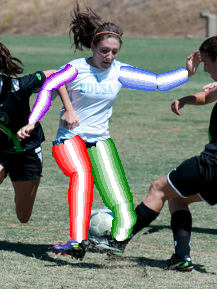}
  \end{subfigure}
  \hfill
    \\
  \begin{subfigure}{0.24\linewidth}    
    \centering
    \includegraphics[height=\height]{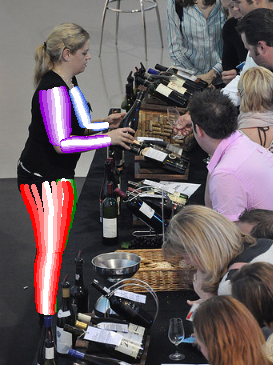}
  \end{subfigure} \vspace{0.1cm}
  \hfill 
  \begin{subfigure}{0.24\linewidth}  
    \centering
    \includegraphics[height=\height]{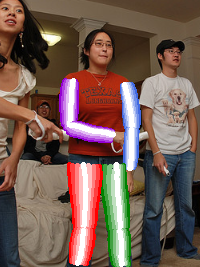}
  \end{subfigure}
  \hfill  
  \begin{subfigure}{0.24\linewidth} 
    \centering
    \includegraphics[height=\height]{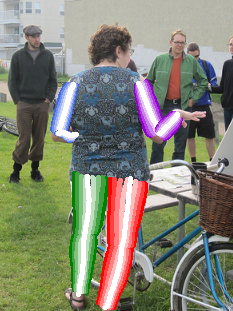}
  \end{subfigure}
  \hfill  
  \begin{subfigure}{0.24\linewidth}  
    \centering
    \includegraphics[height=\height]{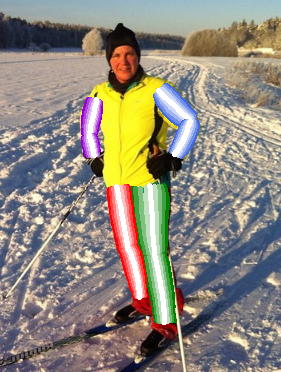}
  \end{subfigure}
  \hfill
  \\
  \begin{subfigure}{0.24\linewidth}    
    \centering
    \includegraphics[height=\height]{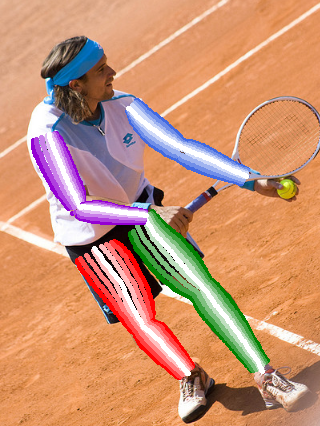}
  \end{subfigure}
  \hfill 
  \begin{subfigure}{0.24\linewidth}  
    \centering
    \includegraphics[height=\height]{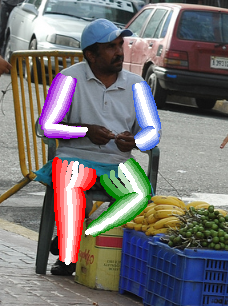}
  \end{subfigure}
  \hfill  
  \begin{subfigure}{0.24\linewidth} 
    \centering
    \includegraphics[height=\height]{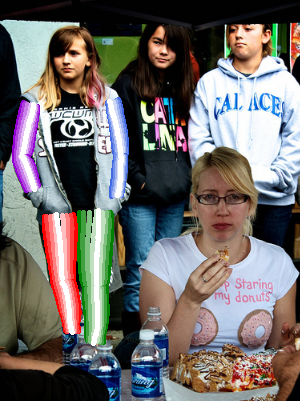}
  \end{subfigure}
  \hfill  
  \begin{subfigure}{0.24\linewidth}  
    \centering
    \includegraphics[height=\height]{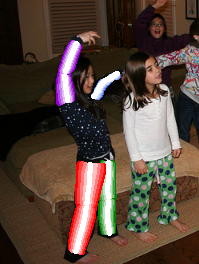}
  \end{subfigure}
  \hfill
  \caption{Examples for model predictions on the DensePose subset of the COCO dataset. The images show four equally spaced lines regarding the thickness on each body part. The projection line is colored white with a color gradient to the edges.}
\end{figure*}

\begin{figure*}[ht]
  \begin{subfigure}{0.24\linewidth}  
  \centering
    \includegraphics[height=\height]{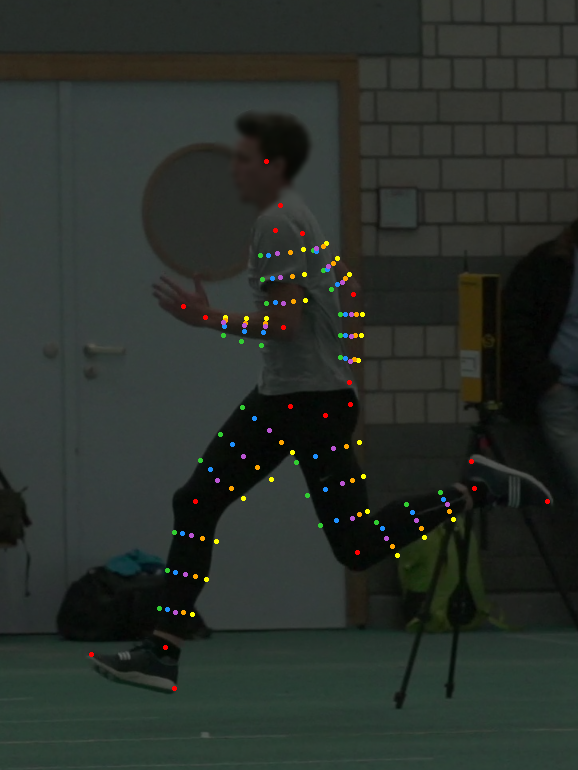}
  \end{subfigure}  \vspace{0.1cm}
  \hfill
  \begin{subfigure}{0.24\linewidth}
    \centering
    \includegraphics[height=\height]{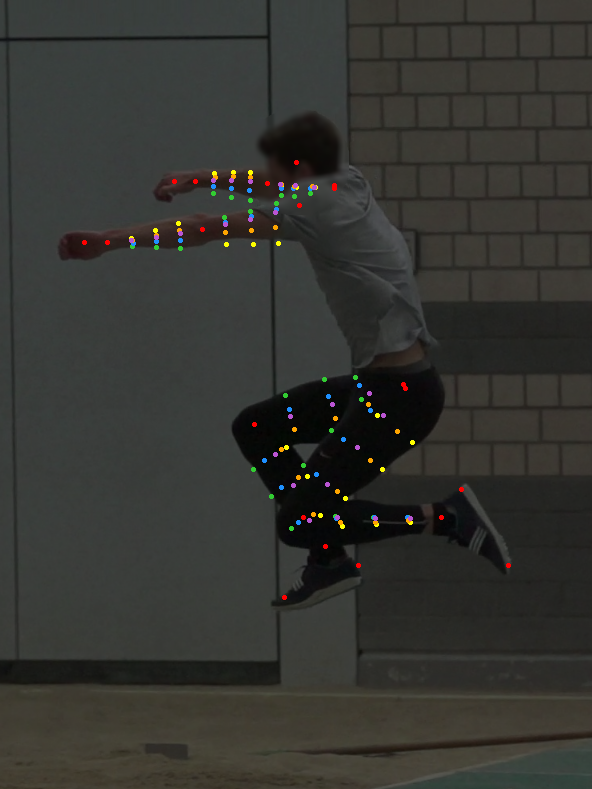}
  \end{subfigure}
  \hfill
  \begin{subfigure}{0.24\linewidth}
    \centering
    \includegraphics[height=\height]{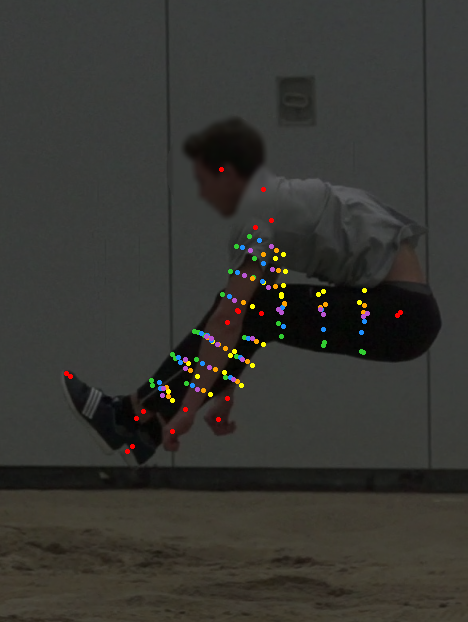}
  \end{subfigure}
  \hfill
  \begin{subfigure}{0.24\linewidth}
    \centering
      \includegraphics[height=\height]{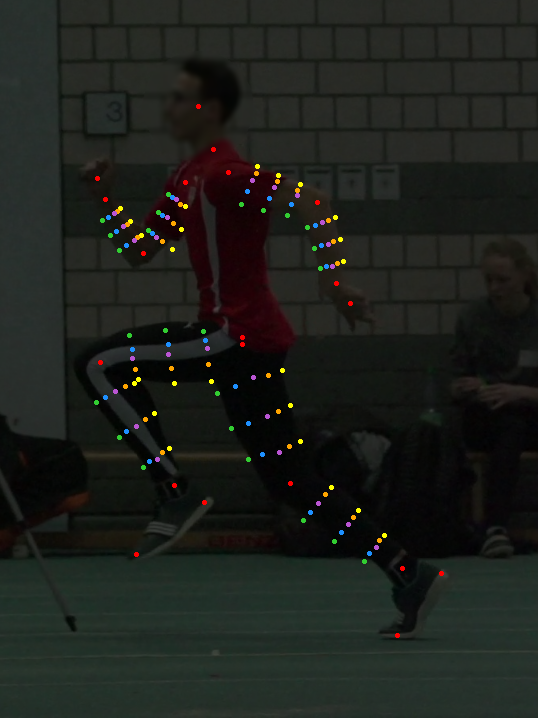}
  \end{subfigure}
  \hfill
  \\
  \begin{subfigure}{0.24\linewidth}    
    \centering
    \includegraphics[height=\height]{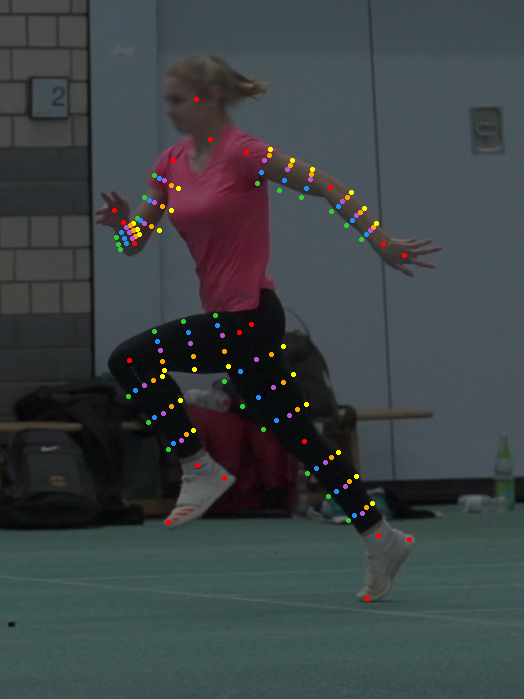}
  \end{subfigure} \vspace{0.1cm}
  \hfill 
  \begin{subfigure}{0.24\linewidth}  
    \centering
    \includegraphics[height=\height]{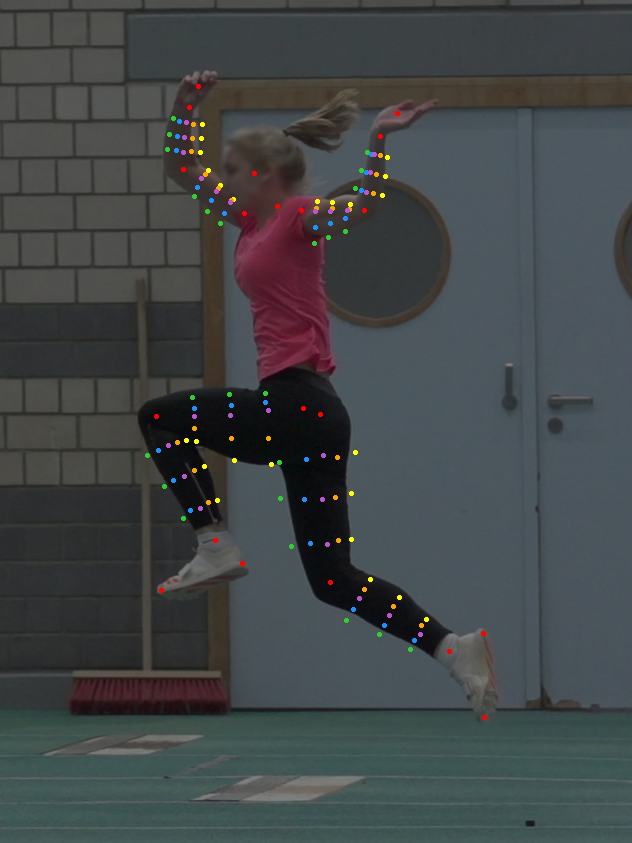}
  \end{subfigure}
  \hfill  
  \begin{subfigure}{0.24\linewidth} 
    \centering
    \includegraphics[height=\height]{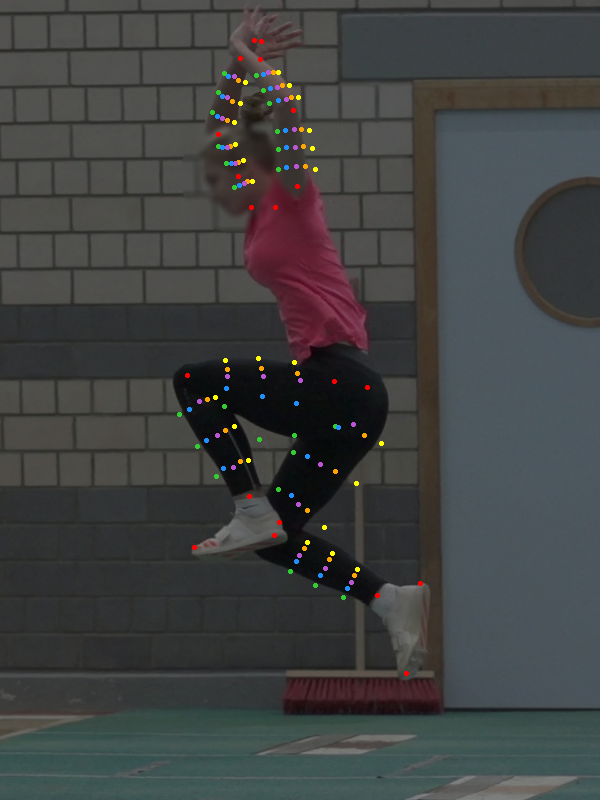}
  \end{subfigure}
  \hfill  
  \begin{subfigure}{0.24\linewidth}  
    \centering
    \includegraphics[height=\height]{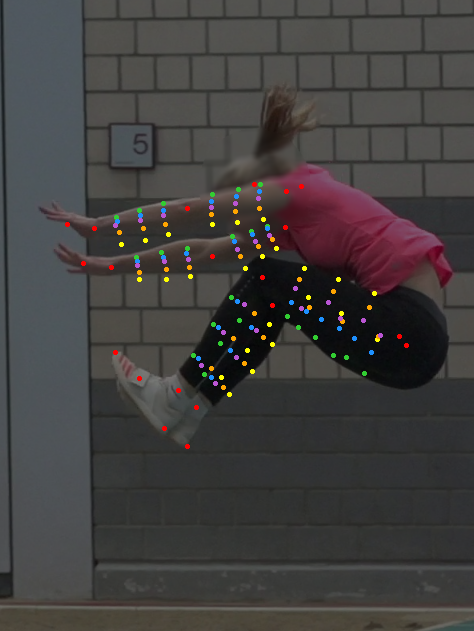}
  \end{subfigure}
  \hfill
    \\
  \begin{subfigure}{0.24\linewidth}    
    \centering
    \includegraphics[height=\height]{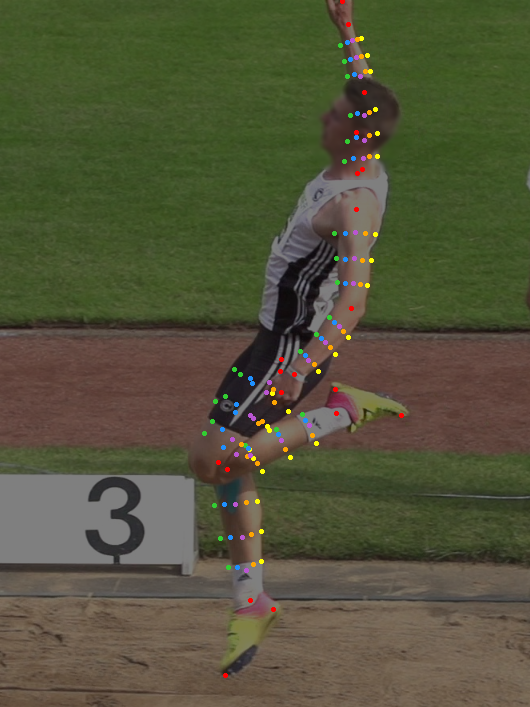}
  \end{subfigure} \vspace{0.1cm}
  \hfill 
  \begin{subfigure}{0.24\linewidth}  
    \centering
    \includegraphics[height=\height]{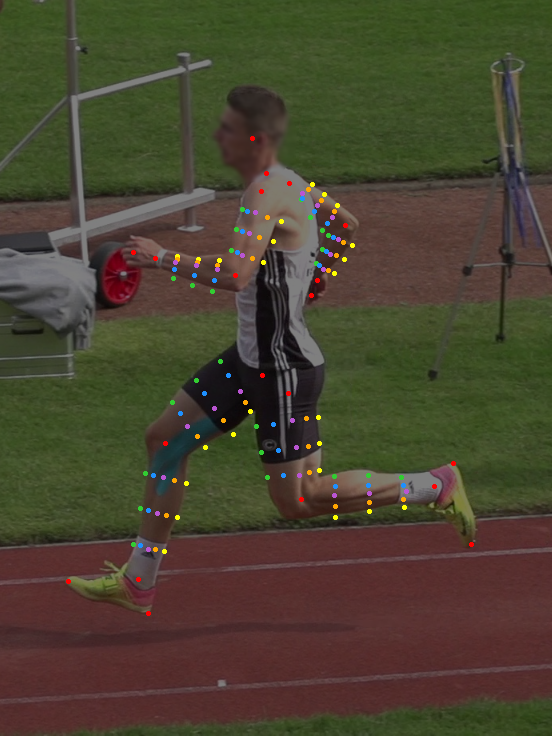}
  \end{subfigure}
  \hfill  
  \begin{subfigure}{0.24\linewidth} 
    \centering
    \includegraphics[height=\height]{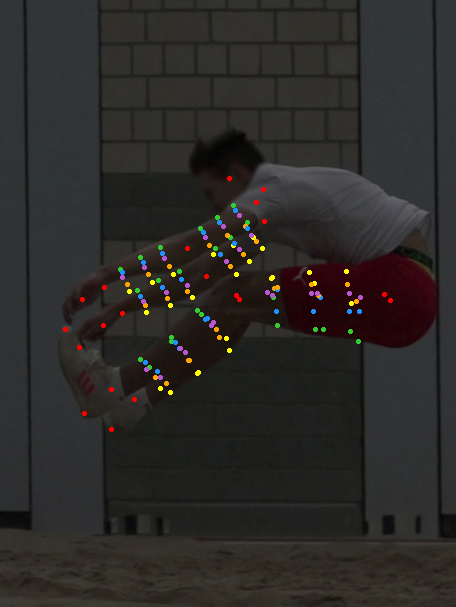}
  \end{subfigure}
  \hfill  
  \begin{subfigure}{0.24\linewidth}  
    \centering
    \includegraphics[height=\height]{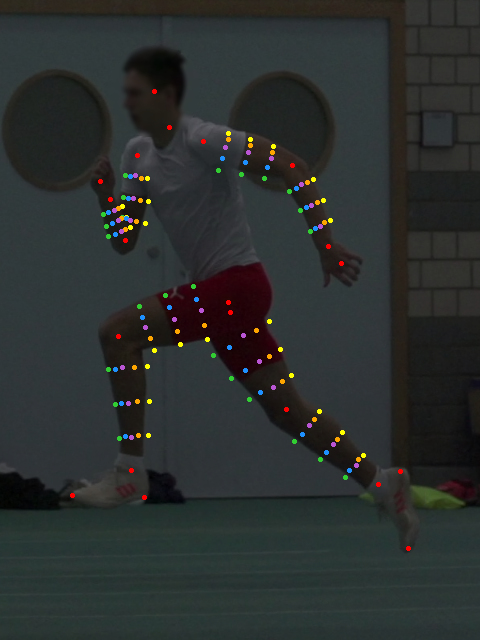}
  \end{subfigure}
  \hfill
  \\
  \begin{subfigure}{0.24\linewidth}    
    \centering
    \includegraphics[height=\height]{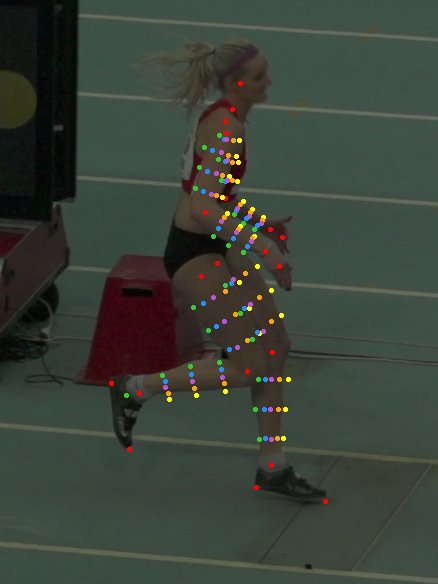}
  \end{subfigure}
  \hfill 
  \begin{subfigure}{0.24\linewidth}  
    \centering
    \includegraphics[height=\height]{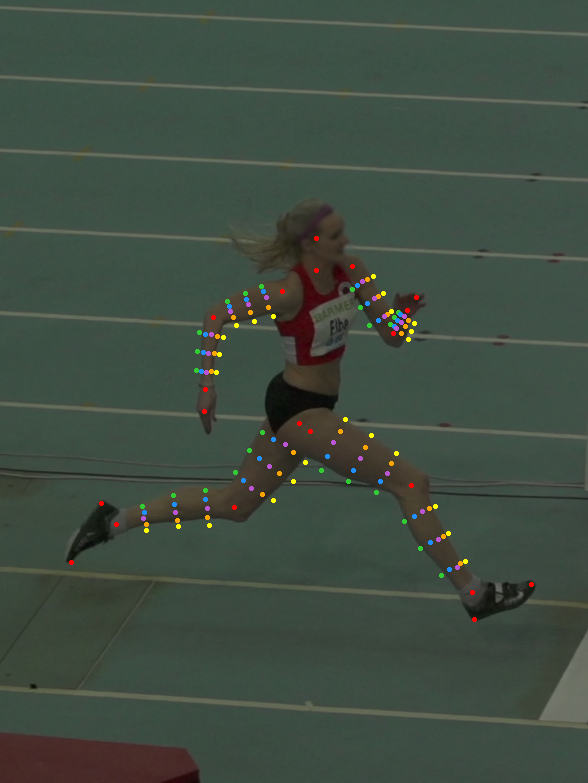}
  \end{subfigure}
  \hfill  
  \begin{subfigure}{0.24\linewidth} 
    \centering
    \includegraphics[height=\height]{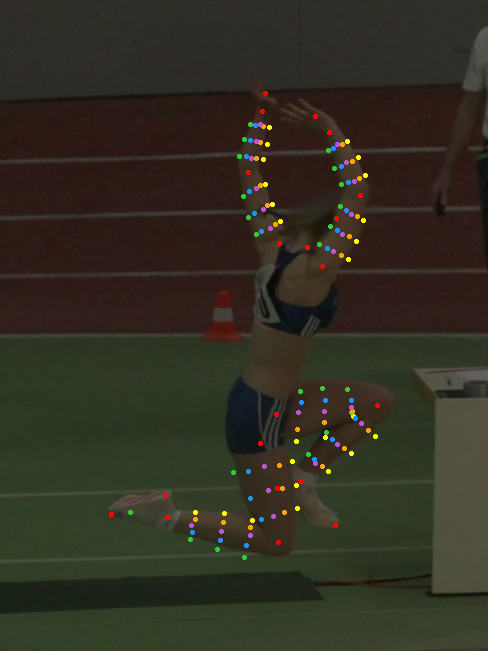}
  \end{subfigure}
  \hfill  
  \begin{subfigure}{0.24\linewidth}  
    \centering
    \includegraphics[height=\height]{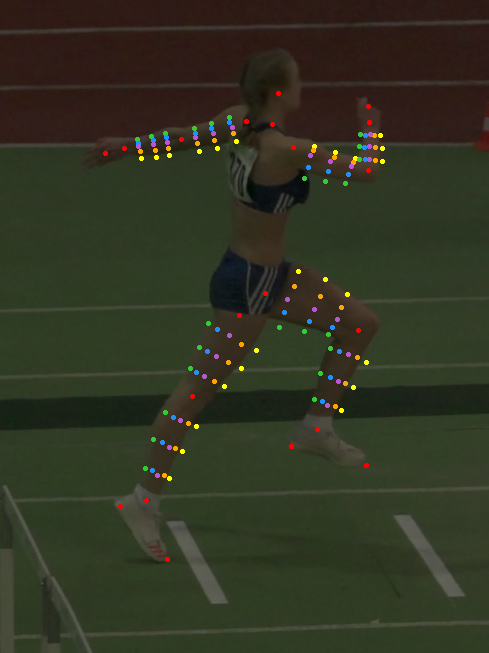}
  \end{subfigure}
  \hfill
  \caption{Examples for model predictions on the triple and long jump dataset including more and less challenging poses. The images show the fixed keypoints in red and a grid of four equally spaced keypoints along the projection line by five equally spaced keypoints along the thickness line for each body part. The images are darkened for better visibility of the keypoints.}
\end{figure*}

\begin{figure*}[ht]
  \begin{subfigure}{0.24\linewidth}  
  \centering
    \includegraphics[height=\height]{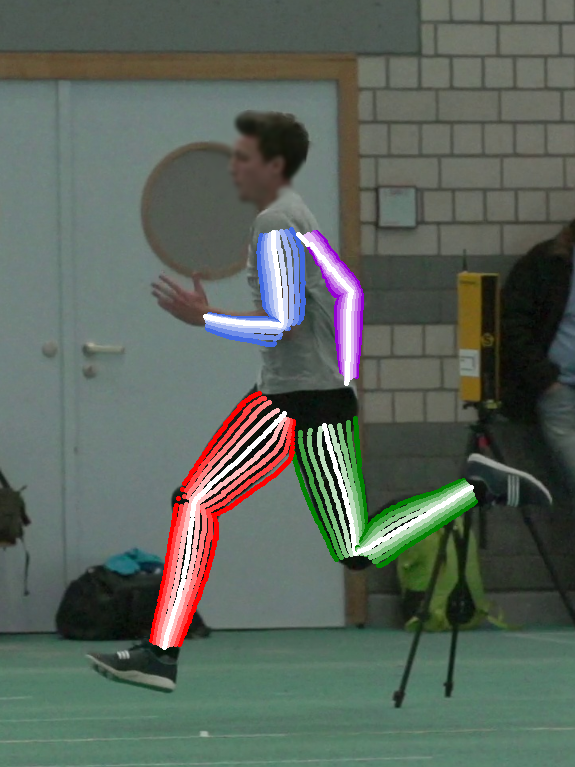}
  \end{subfigure}  \vspace{0.1cm}
  \hfill
  \begin{subfigure}{0.24\linewidth}
    \centering
    \includegraphics[height=\height]{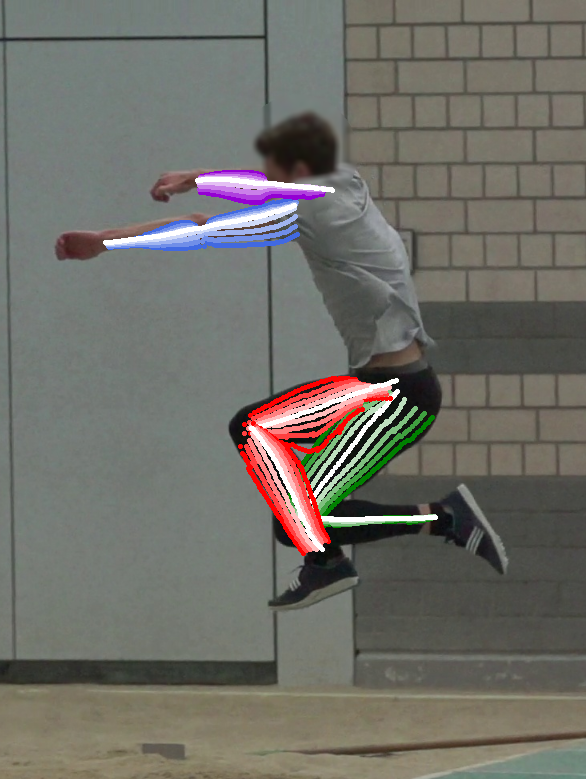}
  \end{subfigure}
  \hfill
  \begin{subfigure}{0.24\linewidth}
    \centering
    \includegraphics[height=\height]{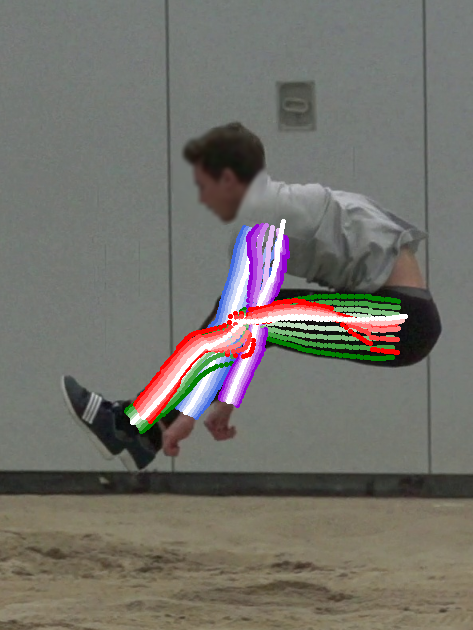}
  \end{subfigure}
  \hfill
  \begin{subfigure}{0.24\linewidth}
    \centering
      \includegraphics[height=\height]{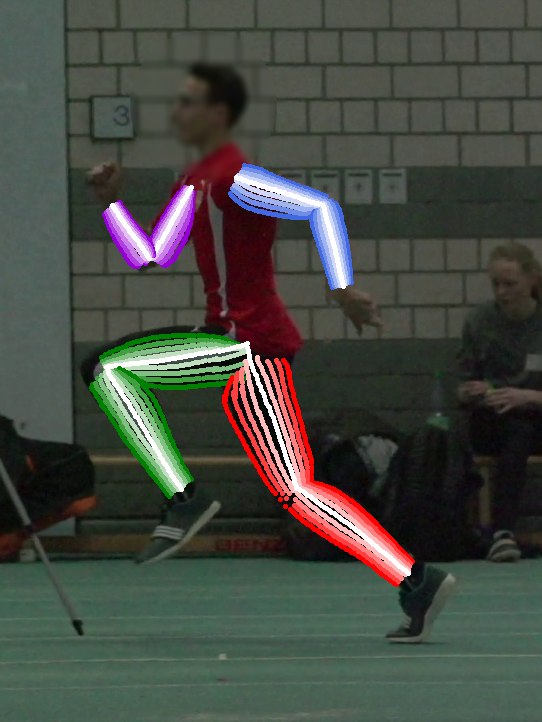}
  \end{subfigure}
  \hfill
  \\
  \begin{subfigure}{0.24\linewidth}    
    \centering
    \includegraphics[height=\height]{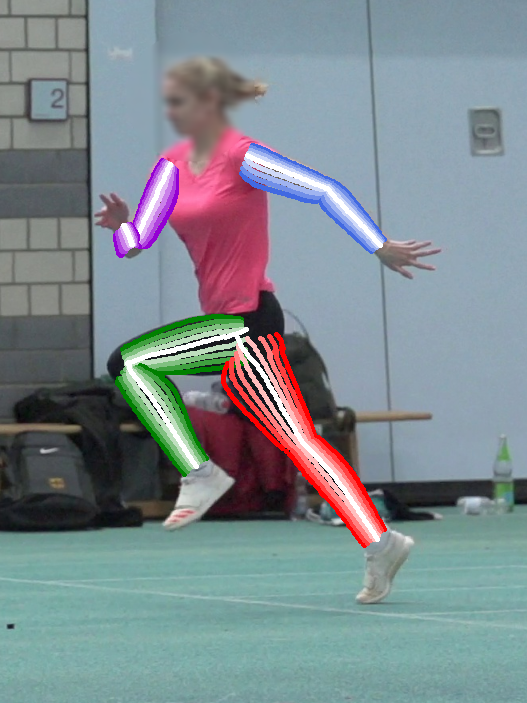}
  \end{subfigure} \vspace{0.1cm}
  \hfill 
  \begin{subfigure}{0.24\linewidth}  
    \centering
    \includegraphics[height=\height]{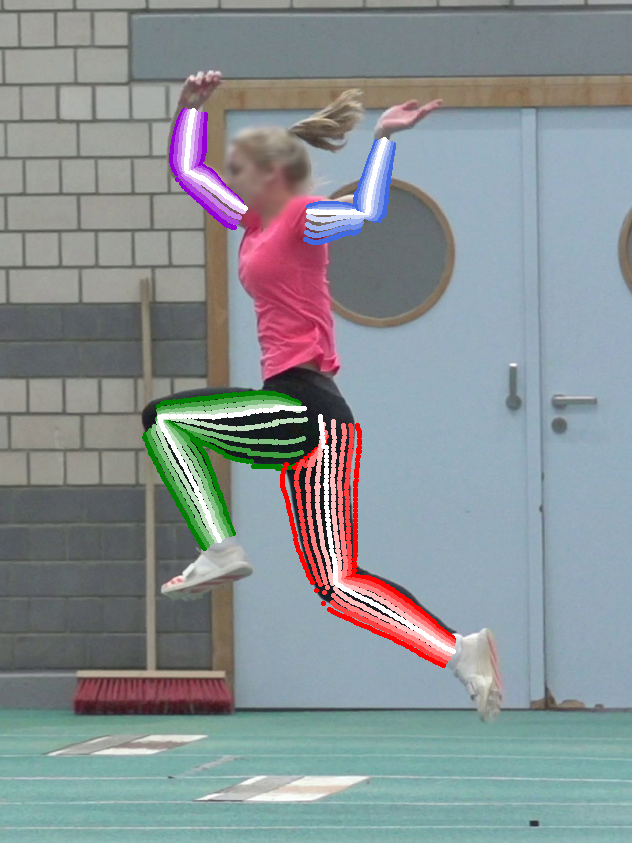}
  \end{subfigure}
  \hfill  
  \begin{subfigure}{0.24\linewidth} 
    \centering
    \includegraphics[height=\height]{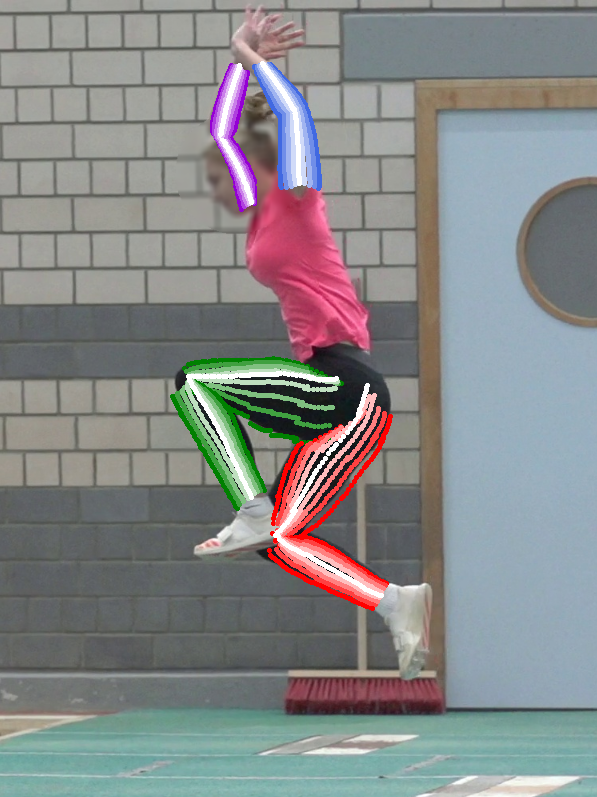}
  \end{subfigure}
  \hfill  
  \begin{subfigure}{0.24\linewidth}  
    \centering
    \includegraphics[height=\height]{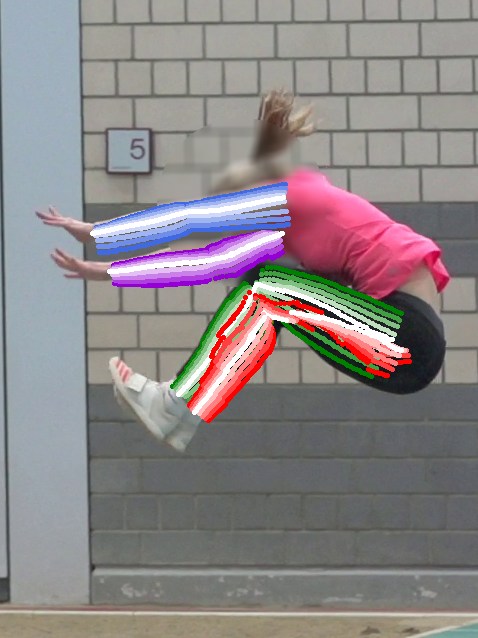}
  \end{subfigure}
  \hfill
    \\
  \begin{subfigure}{0.24\linewidth}    
    \centering
    \includegraphics[height=\height]{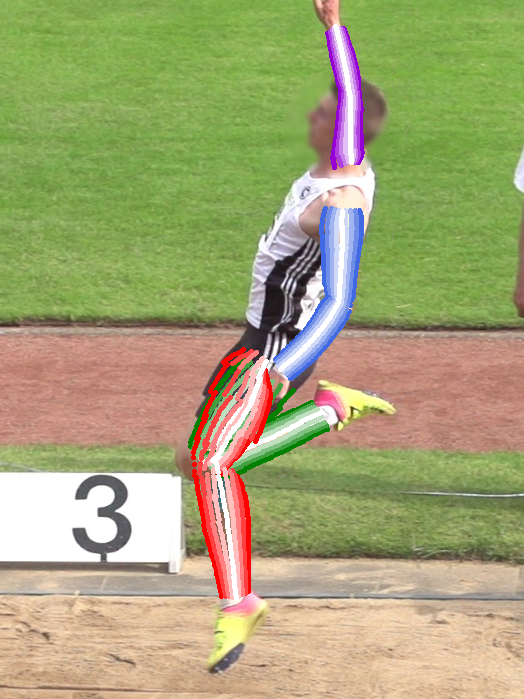}
  \end{subfigure} \vspace{0.1cm}
  \hfill 
  \begin{subfigure}{0.24\linewidth}  
    \centering
    \includegraphics[height=\height]{"line_vis_jump/2"}
  \end{subfigure}
  \hfill  
  \begin{subfigure}{0.24\linewidth} 
    \centering
    \includegraphics[height=\height]{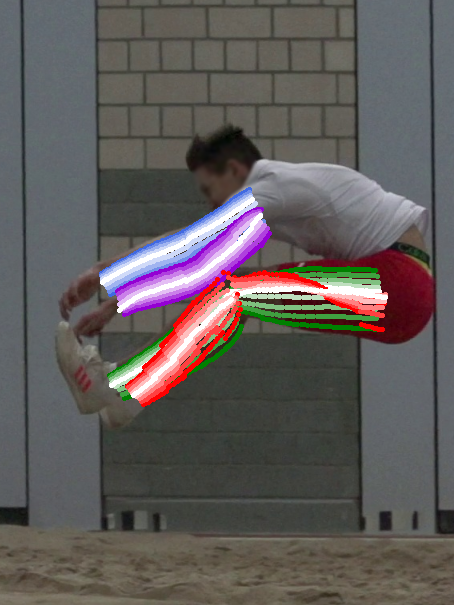}
  \end{subfigure}
  \hfill  
  \begin{subfigure}{0.24\linewidth}  
    \centering
    \includegraphics[height=\height]{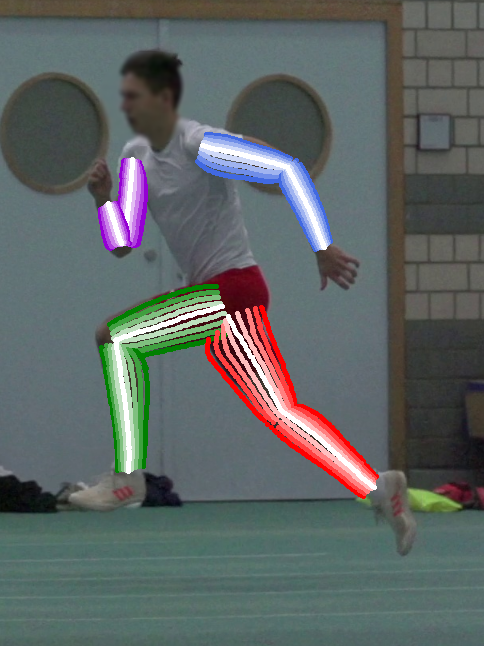}
  \end{subfigure}
  \hfill
  \\
  \begin{subfigure}{0.24\linewidth}    
    \centering
    \includegraphics[height=\height]{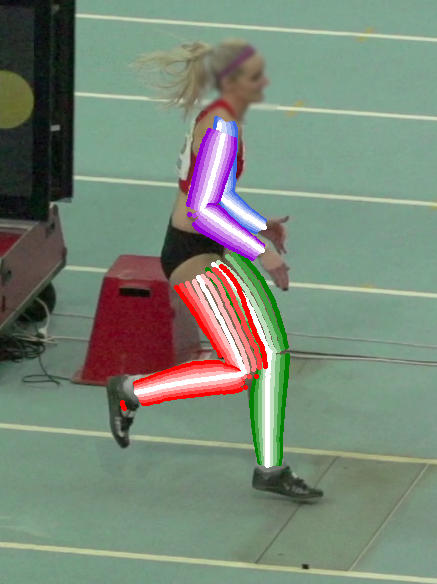}
  \end{subfigure}
  \hfill 
  \begin{subfigure}{0.24\linewidth}  
    \centering
    \includegraphics[height=\height]{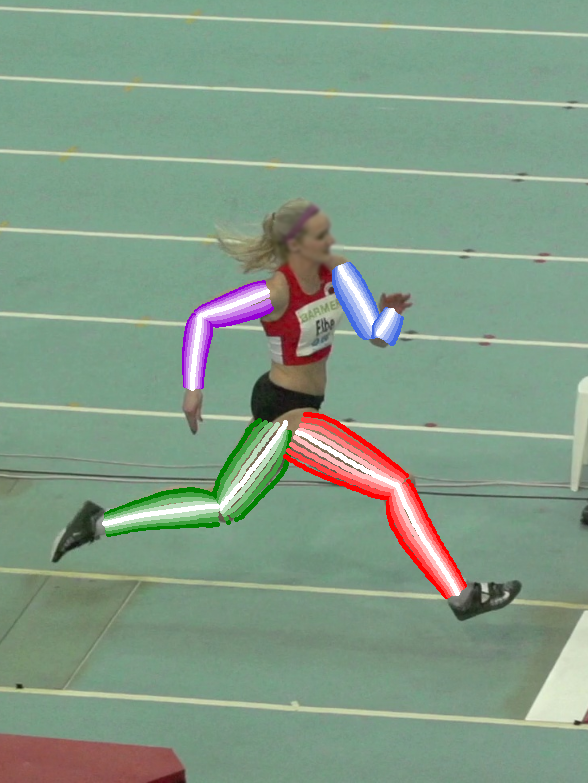}
  \end{subfigure}
  \hfill  
  \begin{subfigure}{0.24\linewidth} 
    \centering
    \includegraphics[height=\height]{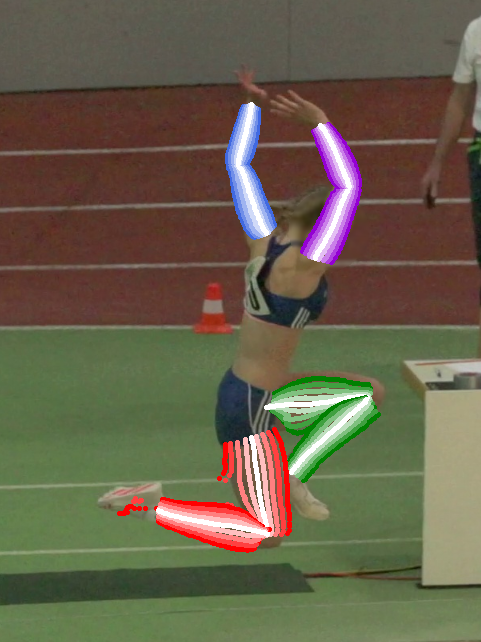}
  \end{subfigure}
  \hfill  
  \begin{subfigure}{0.24\linewidth}  
    \centering
    \includegraphics[height=\height]{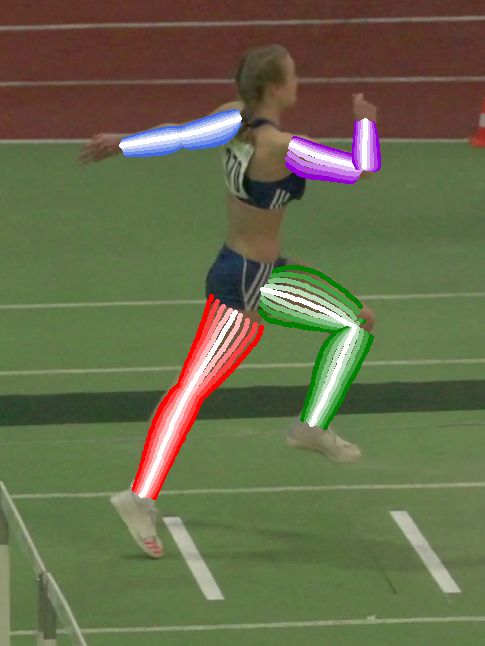}
  \end{subfigure}
  \hfill
  \caption{Examples for model predictions on the triple and long jump dataset including more and less challenging poses. The images show four equally spaced lines regarding the thickness on each body part. The projection line is colored white with a color gradient to the edges.}
\end{figure*}

\end{document}